%% file: main.tex
\pdfoutput=1
\documentclass[10pt,twocolumn,letterpaper]{article}

\usepackage[pagenumbers]{cvpr} % To force page numbers, e.g. for an arXiv version

\usepackage{graphicx}
\usepackage{amsmath}
\usepackage{amssymb}
\usepackage{booktabs}
\input{macros}

\usepackage[pagebackref,breaklinks,colorlinks]{hyperref}

\usepackage[capitalize]{cleveref}
\crefname{section}{Sec.}{Secs.}
\Crefname{section}{Section}{Sections}
\Crefname{table}{Table}{Tables}
\crefname{table}{Tab.}{Tabs.}

\def\cvprPaperID{4592} % *** Enter the CVPR Paper ID here
\def\confName{CVPR}
\def\confYear{2023}

\begin{document}

%%%%%%%%% TITLE
\title{Internal Diverse Image Completion}

\author{
Noa Alkobi\\
Technion\\
\and
Tamar Rott Shaham\\
Technion, MIT\
\and
Tomer Michaeli\\
Technion\\
}

% \and

% \author{%
%   Noa Alkobi\\
%   Technion–Israel Institute of Technology\\
%   \texttt{noalkobi@gmail.com} \\
%   % examples of more authors
%   \and
%   Tamar Rott Shaham \\
%   Technion–Israel Institute of Technology\\
%   \texttt{stamarot@campus.technion.ac.il} \\
%   \and
%   Tomer Michaeli\\
%   Technion–Israel Institute of Technology\\
%   \texttt{tomer.m@ee.technion.ac.il}
% }

% \author{Noa Alkobi\\
% Technion\\
% \and
% Tamar Rott Shaham\\
% Technion, MIT\
% \and
% Tomer Michaeli\\
% Technion\\
% }

\maketitle
\thispagestyle{empty}

\input{sections/0_abstract}

\input{sections/1_intro}
\input{sections/2_related}
\input{sections/3_method}

\input{sections/4_experiments}

\input{sections/5_conclusions}

%\clearpage\mbox{}Page \thepage\ of the manuscript.
%\clearpage\mbox{}Page \thepage\ of the manuscript.

\clearpage
\bibliographystyle{ieee_fullname}
\bibliography{bib}
\end{document}

\title{Internal Diverse Image Completion \\ Supplementary Material}
\author{Noa Alkobi\\
Technion\\
\and
Tamar Rott Shaham\\
Technion, MIT\
\and
Tomer Michaeli\\
Technion\\
}

% \and

\maketitle
% \renewcommand{\cftdot}{} %empty {} for no dots. you can have any symbol inside. For example put {\ensuremath{\ast}} and see what happens.
% \tableofcontents

% \addtocontents{toc}{\protect\thispagestyle{empty}}
\tableofcontents
% \pagenumbering{arabic}
% \pagenumbering{}
% \AtBeginDocument{\addtocontents{toc}}
\clearpage
\thispagestyle{empty}

% \documentclass[runningheads]{llncs}
% \usepackage{graphicx}
% % DO NOT USE \usepackage{times}, it will be removed by typesetters
% %\usepackage{times}

% \usepackage{tikz}
% \usepackage{comment}
% \usepackage{amsmath,amssymb} % define this before the line numbering.
% \usepackage{color}
% \usepackage{xcolor}
% \usepackage[export]{adjustbox}
% \usepackage[subrefformat=parens]{subcaption}

% \usepackage{multirow}
% \usepackage{booktabs}

% % INITIAL SUBMISSION - The following two lines are NOT commented
% % CAMERA READY - Comment OUT the following two lines
% \usepackage{ruler}
% \usepackage[width=122mm,left=12mm,paperwidth=146mm,height=193mm,top=12mm,paperheight=217mm]{geometry}

% \input{macros}
\renewcommand{\thefigure}{S\arabic{figure}}
\renewcommand{\thetable}{S\arabic{table}}
% \begin{document}
% % \renewcommand\thelinenumber{\color[rgb]{0.2,0.5,0.8}\normalfont\sffamily\scriptsize\arabic{linenumber}\color[rgb]{0,0,0}}
% % \renewcommand\makeLineNumber {\hss\thelinenumber\ \hspace{6mm} \rlap{\hskip\textwidth\ \hspace{6.5mm}\thelinenumber}}
% % \linenumbers
% \pagestyle{headings}
% \mainmatter
% \def\ECCVSubNumber{3109}  % Insert your submission number here

% \title{Internal Diverse Image Completion \\ Supplementary Material} % Replace with your title

% % INITIAL SUBMISSION 
% %\begin{comment}
% \titlerunning{ECCV-22 submission ID \ECCVSubNumber} 
% \authorrunning{ECCV-22 submission ID \ECCVSubNumber} 
% \author{Anonymous ECCV submission}
% \institute{Paper ID \ECCVSubNumber}
% %\end{comment}
% %******************

% % CAMERA READY SUBMISSION
% \begin{comment}
% % If the paper title is too long for the running head, you can set
% % an abbreviated paper title here
% \end{comment}
% %******************
% \maketitle

\input{sections/6_supplementary}

\clearpage
\bibliographystyle{ieee_fullname}
\bibliography{bib}
\clearpage

%% file: macros.tex
\usepackage{amsmath,amsfonts,bm,relsize,dsfont}
\usepackage{color}
\usepackage{xcolor}
\usepackage{wrapfig}

\newcommand{\myparagraph}[1]{\vspace{1pt}\noindent\textbf{#1}}

\def\inputImg{y}
\def\real{x} % input
\def\noise{z} % input

\def\noiserec{z^{\text{rec}}} % input
\def\fake{\tilde{x}} % output

\def\mask{m}

%% file: sections/0_abstract.tex
\begin{abstract}
\input{figures/figure1}

Image completion is widely used in photo restoration and editing applications, \eg for object removal. Recently, there has been a surge of research on generating diverse completions for missing regions. However, existing methods require large training sets from a specific domain of interest, and often fail on general-content images. In this paper, we propose a diverse completion method that does not require a training set and can thus treat arbitrary images from any domain. Our internal diverse completion (IDC) approach draws inspiration from recent single-image generative models that are trained on multiple scales of a single image, adapting them to the extreme setting in which only a small portion of the image is available for training. We illustrate the strength of IDC on several datasets, using both user studies and quantitative comparisons.
%We propose a novel Diverse Internal Completion method. A model trained on a partial single image which generates diverse and realistic results for the inpainting task. Our model is based on a pyramid of generators, patch discriminators and the masked image in different sizes from coarse to fine scales. At coarse scales the generator learns to generate the global structure of the image, and when proceeding in training the fine generators learn to generate finer details in the image. 
%At inference, we offer the option of sampling results with varying diversity, ranging from low diversity and not changing the knowing pixels in the image to higher diversity and allowing to change pixels next to the missing pixels in the original image.
%Our new method was tested on Places2 and Part-Imagenet datasets and showed diversity and realism in the results.
\end{abstract}

%% file: figures/figure1.tex
\begin{figure*}[t]
\includegraphics[width=1.8\columnwidth]{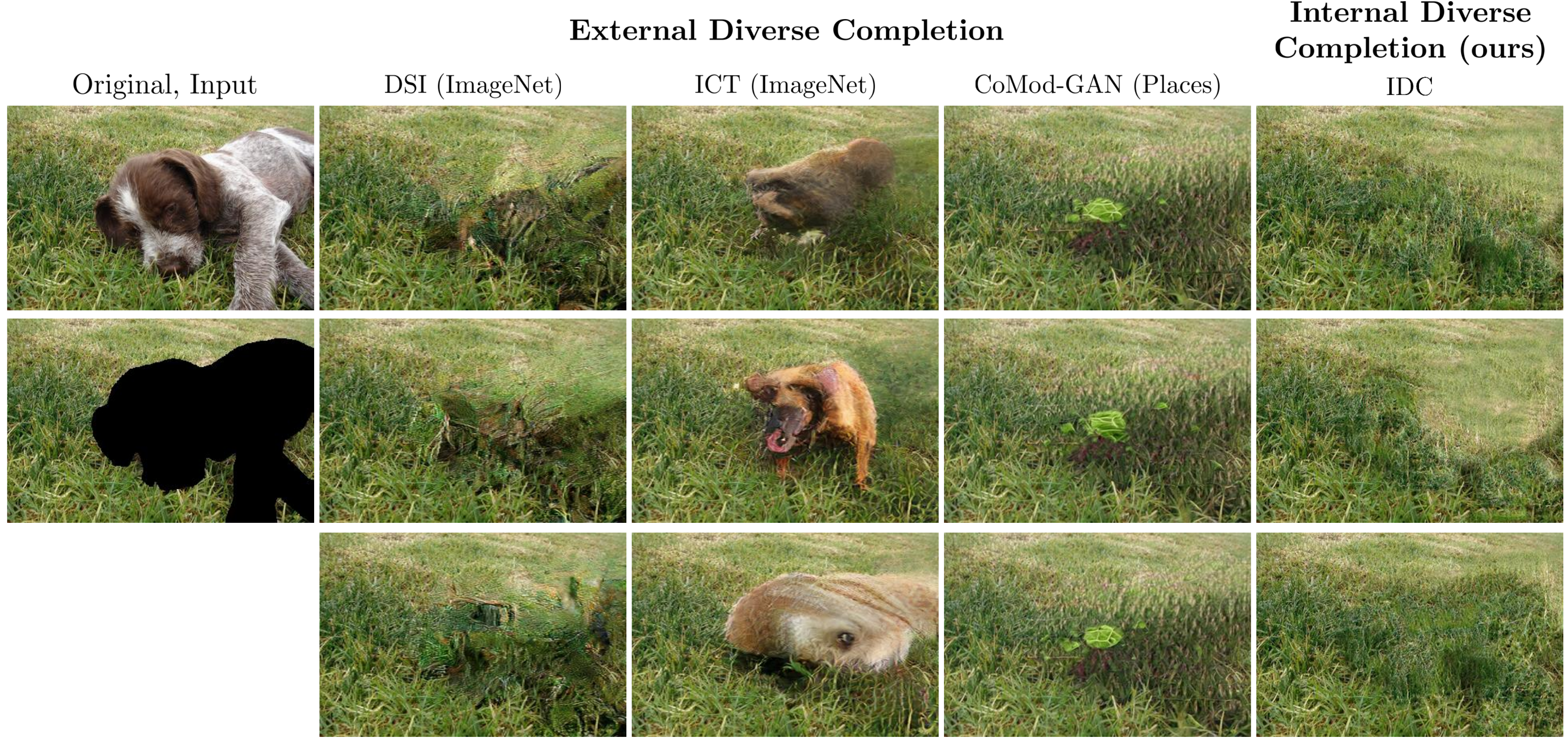}
\centering
	\caption{\label{fig:figure1}
	\textbf{Internal Diverse Completion.} We introduce a new single image GAN model targeted for the task of image completion. Our model is trained on a single image with a missing region, and then suggests diverse completions based on only the internal statistics of the image to be inpainted itself. In the context of object removal, externally trained models are often unsuitable, as they attempt to insert alternative objects in the missing region (similar faliures occure with DSI and ICT models trained on Places; see supplementary).\vspace{-0.5mm}
	%\textbf{Examples of our Diverse Internal Inpainting (DII) method.} From top to bottom we show the GT image, the masked input image and several samples. From left to right we show results with high diversity, medium diversity and low diversity compared with the mask size.
	}
\end{figure*}

%% file: sections/1_intro.tex
\section{Introduction}

Image completion (or inpainting) refers to the problem of filling-in a missing region within an image. This task is of wide applicability in image restoration, editing and manipulation applications (\eg for object removal), and has thus seen intense research efforts over the years.  
%This is a challenging task since the missing pixels need to match with the known pixels of the image and be unrecognized to the human eye.
%Inpainting is a basic problem in the field of image processing, which can be used for image editing, object removal and more, thus has been explored for decades.
Since the first inpainting method of Bertalmio et al.~\cite{bertalmio2000image}, image completion techniques have rapidly evolved from methods that locally diffuse information into the missing region \cite{ballester2001filling,bertalmio2003simultaneous,telea2004image}, to algorithms that copy chunks from other locations within the image \cite{pritch2009shift,barnes2009patchmatch,xu2010image}, and recently, to deep learning approaches that are trained on large external datasets \cite{yu2018generative,qu2021structure,xie2012image,pathak2016context}.

The transition to externally trained models has enabled the treatment of challenging inpainting settings, like completion of large parts of semantic objects, and \emph{diverse inpainting} \cite{zheng2019pluralistic,peng2021generating,zhao2020uctgan,cai2019diversity}, \ie generating a variety of different completions. However, despite their merits, externally trained methods are limited by the need for very large training sets from the particular domain of interest (\eg faces \cite{liu2015faceattributes,CelebAMask-HQ}, bedrooms \cite{yu2015lsun}, etc.). Even when trained on more generic datasets, like Places~\cite{zhou2014learning} or Imagenet~\cite{deng2009imagenet}, such models are still often limited in the object categories or photograph styles they can handle. An additional limitation of external models is that they cannot be trained on the task of object removal (as opposed to generic inpainting), because this requires pairs of images, one depicting an object over some background, and one depicting only the background behind the object. Such pairs are obviously unavailable in real-world scenarios, but seem to be crucial for successful object removal. Indeed, when generic external inpainting methods are used for object removal, they tend to insert alternative objects instead of filling in the missing region with background. This is illustrated in Fig.~\ref{fig:figure1}. In light of those limitations, it is natural to ask whether external training is the only route towards generic diverse completion.

In this paper, we explore the idea of \emph{internal diverse completion} (IDC). We propose an inpainting method that is capable of generating a plethora of different completions, without being exposed to any external training example \footnote{Code is available at: \url{https://github.com/NoaAlkobi/IDC}}. 
IDC can be applied to images from any domain and is not constrained to particular image dimensions, aspect ratios, or mask shapes and locations. The reliance on internal learning, obviously limits IDC in its ability to complete large parts of semantic objects (\eg parts of a face or an animal). However, in object removal scenarios, this is typically not required. We illustrate that in such cases, IDC performs at least on par with externally trained models when applied to images from the domain on which they were trained, while it can also be successfully applied to other domains. One representative within-domain example is shown in Fig.~\ref{fig:figure1}, where the external DSI~\cite{peng2021generating}, ICT~\cite{wan2021high}, and CoMod-GAN~\cite{zhao2021large} models do not provide plausible completions. This is while IDC manages to generate diverse photo-realistic completions for the grass. Completion examples for other domains are shown in Fig.~\ref{fig:figure2}. 

%it is natural the ask whether external supervision is necessary for obtaining diverse completion.

%Image completion is an ill posed task, in the sense that the missing region can typically be completed in many different plausible ways. 

%Several recent works presented methods for diverse inpainting \cite{zheng2019pluralistic,peng2021generating,zhao2020uctgan,cai2019diversity}. Those models were trained on a huge dataset and although they produce diverse and reasonable results they might fail when given an input which does not come from the same distribution as the training dataset.

%Despite this fact, most methods until recently, provided only a single solution. Lately, there have been several papers that suggested methods for external diverse inpainting \cite{zheng2019pluralistic,peng2021generating,zhao2020uctgan,cai2019diversity}. Those models were trained on a huge dataset and although they produce diverse and reasonable results they might fail when given an input which does not come from the same distribution as the training dataset.

%Inpainting can also be done with a model trained on a single image as proposed in \cite{ulyanov2018deep}, but the result is a single low resolution image.

%The question being asked is whether a model, trained on a partial single image, can learn to generate diverse and realistic solutions for the inpainting task?
%Surprisingly, in this paper we demonstrate a new method for Diverse Internal Image Completion (IDC).
We make use of recent progress in single-image generative models. Particularly, we adopt the framework of SinGAN \cite{shaham2019singan}, which is a multi-scale patch-GAN model that can be trained on a single natural image. However, our problem is more challenging than that treated in \cite{shaham2019singan} because (i) only a small part of the image is available for training, especially at the coarser scales, and (ii) the generated content must match and seamlessly blend with the neighboring regions. We address these difficulties through a dedicated architecture, and training and inference processes.

%one partial real image and is able to generate diverse plausible results for the missing area in the image.

%Our contributions are summarized as follows:
%\begin{itemize}
%    \item Based on a pyramid of GANs, we introduce a novel method for Diverse Internal Completion (IDC).
%    \item We show that our approach generates reasonable results and can even be compared with external diverse inpainting methods.
%    \item We propose the option to control the diversity of the results against preserving the known pixels in the masked image.
%\end{itemize}

%% file: sections/2_related.tex
\section{Related work}

\paragraph{Internal inpainting} Internal completion methods rely solely on the information within the image to be inpainted. Methods in this category include diffusion-based approaches~\cite{ballester2001filling,levin2003learning,telea2004image,bertalmio2001navier} and patch-based algorithms~\cite{barnes2009patchmatch,criminisi2004region,xu2010image,bertalmio2003simultaneous,drori2003fragment,simakov2008summarizing,pritch2009shift}. %approaches fill the missing area in the image by propagating information from the background of the image.
% Learning the internal patch distribution within a single natural image has been extensively investigated over the last years \cite{zontak2011internal}. Lately, many works exploited the fact that patches in an image tend to recur in many locations and scales within the same image for different tasks.
% Particularly, learning the internal distribution of single images was recognized as a powerful prior for super resolution \cite{glasner2009super}, denoising \cite{zontak2013separating}, deblurring \cite{michaeli2014blind}, dehazing \cite{bahat2016blind} and more.
Recently, it was shown that internal inpainting can also be achieved by overfitting a deep neural network (DNN) to the input image%using the internal information within a single image was taken into the deep learning regime. 
~\cite{ulyanov2018deep}. %presents a deep model trained on a single image for image restoration tasks, including inpainting.
%, where several single image GAN schemes schemes were presented, such as InGAN \cite{shocher2018ingan} and SinGAN \cite{shaham2019singan}. %Learning the internal distribution of single images was recognized as a powerful prior for several tasks such as inpainting. \cite{ulyanov2018deep} were the first to present a deep model trained on a single image for the task of inpainting, which aims to "overfit" to the specific image. 
%However, although  %the task of 
%inpainting is an ill-posed task,
However, despite the inherent ambiguity in the inpainting task, all internal methods to date are designed to output only a single completion. 
Here, we introduce an internal completion method that can generate diverse completion suggestions. 
%Our model is also internal as we train our model to leverage the information within a single input image, yet it  %method we 
%offers many different %several 
%completions solutions.

\paragraph{External inpainting}
A dramatic leap in the ability to complete semantic contents, was achieved by transitioning to DNNs trained on large datasets. Example approaches include GAN based techniques~\cite{li2017generative,iizuka2017globally,yeh2016semantic} and encoder-decoder frameworks~\cite{liu2020rethinking,pathak2016context,liu2019coherent}% have shown a tremendous improvement in the task of inpainting
. Progress in this route has been driven by the developments of various mechanisms,  %A lot of methods were proposed in order to improve the results of the inpainting 
% including gated and partial convolutions \cite{yu2019free,liu2018image},
including gated, partial and fast Fourier convolutions~\cite{yu2019free,liu2018image,suvorov2022resolution},
%to reduce normal convolutions artifacts in the results of inpainting. \cite{qu2021structure} proposed to use 
multi-scale architectures~\cite{qu2021structure,yang2017high}, %pyramid of GANs and \cite{yang2017high} uses a multi-scale neural patch synthesis for the task of image inpainting. 
edge map representations% as structural guidance
~\cite{nazeri2019edgeconnect,xu2020e2i,ren2019structureflow} %for image inpainting and \cite{ren2019structureflow} suggested to use edge-preserved smooth images instead. 
and contextual attention~\cite{yu2018generative}. %and patch swap \cite{pritch2009shift} to improve results. However, this 
All these methods require large training sets, and like their internal counterparts, output %only 
a single solution. By contrast, our method is trained only on the masked input image %a single image 
and generates diverse completions. %for a specific inpainted image.

\paragraph{Diverse external inpainting} Several recent works developed diverse inpainting methods based on VAEs~\cite{zhao2020uctgan,zheng2019pluralistic,peng2021generating}, 
%Generative Adversarial Networks (
GANs~% with several modifications 
\cite{liu2021pd,cai2020piigan,zhao2021large},  %In addition
transformers~\cite{wan2021high,yu2021diverse}, and diffusion models~\cite{saharia2021palette}. %have been used to solve the task of diverse image inpainting. 
%However %, in contrast to our method, 
These methods work well on inputs resembling their training data (in content and mask shapes). However, their performance degrades on out-of-distribution inputs. %,  as their similar to their training set distribution, and when
%are adapted to the dataset they were trained on. For example, \cite{han2019finet} which was trained on a fashion dataset will not be able to inpaint images with other distribution, e.g. natural images. In 
Our method %, we 
does not require a training set and is thus not restricted to specific types of images or masks.

\input{figures/figure2}

%\myparagraph{Single image inpainting}
%\todo{This paragraph covers classic methods}

%\myparagraph{Deep Internal learning}
%\todo{This paragraph covers networks trained on a single example (e.g. DIP, ZSSR, SinGAN).}
%\tamar{maybe one paragraph for both single image inpainting and internal learning}.

%% file: figures/figure2.tex
\begin{figure*}[ht]
\includegraphics[width=1.8\columnwidth]{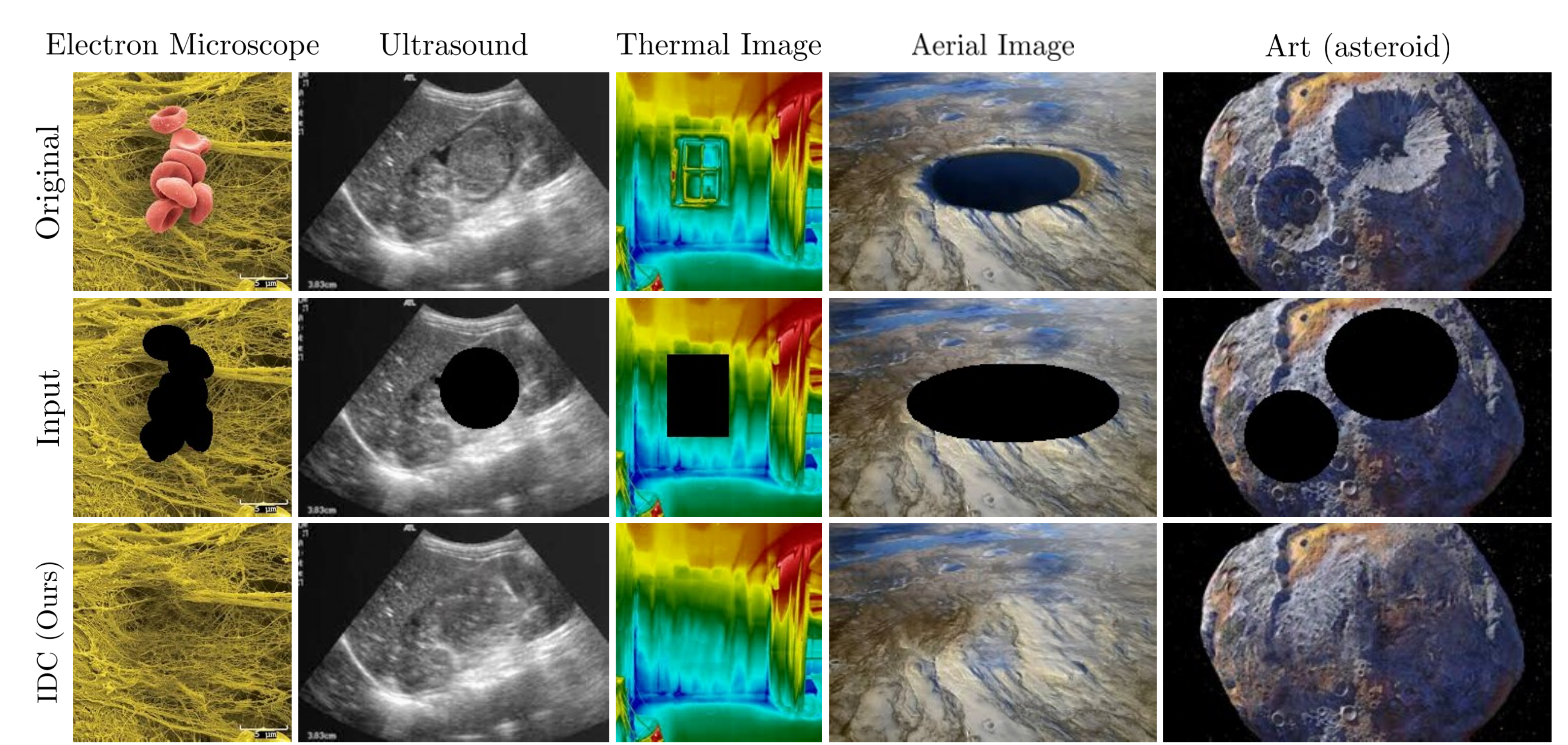}
\centering
	\caption{\label{fig:figure2}
	\textbf{Internal completion for different domains.} Our method can be applied to images from arbitrary domains and is not restricted to a specific mask shape or location. From left to right: blood cells imaged by an electron microscope, ultrasound image of a Wilms tumor, thermal image of a room, aerial photograph of Northern Quebec's Pingualuit Crater, and an artist's conception of the asteroid 16 Psyche. Completions with baselines are provided in SM.
	}
\end{figure*}

%% file: sections/3_method.tex
\section{Method}

We are given a masked image 
\begin{align}
\inputImg=\real\odot(1-\mask),
\end{align}
where $\real$ is the original (non-masked) image, $\mask$ is a binary mask containing $1$'s in the masked regions and $0$'s elsewhere, and $\odot$ denotes per-pixel product. Our goal is to randomly generate estimates $\fake$ of $\real$ that satisfy two properties: (i)~each $\fake$ matches $x$ in the non-masked regions, and (ii) the distribution of patches within $\fake$ is similar to that of the non-masked patches of $\inputImg$.

%We consider the task of image inpainting. Given an 
%Consider a real 
%image $\real$ with missing pixels (indicated by a binary mask \Mask), %(where one denotes the missing pixels and zero the known ones). 
%our goal is to suggest many diverse %several compatible 
%solutions that realistically fill that area. %to the missing pixels in the image.
%To achieve this we train a GAN model that captures the internal patch distribution of the valid area of the image $\real\odot(1-\mask)$ (where $\odot$ represents a pixel-wise product). At inference, our model is able to generate many different realistic completions to the missing area $\real\odot\mask$.

%Our model is 
Our approach is inspired by %from 
SinGAN~\cite{shaham2019singan}, which is a multi-scale patch-GAN architecture that can be trained on a single image. Specifically, our goal is to train a model that captures the distribution of small patches within the valid regions of $y$, at multiple scales, and then use this model to generate content within the masked region. 
To this end, we construct pyramids $\{\inputImg_0,\ldots,\inputImg_N\}$ and $\{\mask_0,\ldots,\mask_N\}$, from the input image $y$ and the binary mask $m$, where $\inputImg_0=\inputImg$, $\mask_0=\mask$, and the scale factor between consecutive levels is $\alpha$.  %which uses a single real image to we train 
Using these images, we train a pyramid of GANs, as shown in Fig.~\ref{fig:architecture}. Specifically, in each pyramid level $n$, we train a generator $G_n$ against a patch discriminator $D_n$~\cite{li2016precomputed,isola2017image}. 
%\todo{cite markovian GAN and pix2pix} 
The generator consumes a white Gaussian noise map, $\noise_n$, having the same dimensions as the input image at that scale, as well as an up-sampled version of the fake image generated by the previous scale (the coarsest scale accepts only noise), and it outputs a fake image $\fake_n$. The discriminator $D_n$ is presented with either the fake image $\fake_n$ or the masked image $\inputImg_n$ (or a processed version of $\inputImg_n$, see Sec.~\ref{sec:training}), and attempts to classify each of the overlapping (valid) patches of its input as real or fake. Thus, its output is a discrimination map.
\input{figures/architecture}

All the generators and discriminators comprise five convolutional blocks consisting of \texttt{conv-BN-LeakyRelU}. Therefore, each GAN in the pyramid captures the distribution of $11\times11$ patches (the networks' receptive field). At the coarser scales, these patches cover large structures relative to the image dimensions, whereas the finer scales capture smaller structures and textures.

%(see SM for ablation of variance of injected noise)

\subsection{Training}\label{sec:training}
Training is done sequentially from the coarsest scale to the finest one. Each scale is trained individually, while keeping the generators of all coarser scales fixed. We optimize an objective function comprising an adversarial loss and a reconstruction loss.

For the \emph{adversarial term}, we use the Wasserstein GAN loss, together with gradient penalty~\cite{gulrajani2017improved}.
%\todo{cite WGAN-GP}. 
%The discrimination score used in this term, is the mean of the discrimination map over all elements belonging to valid patches. More specifically,
For this loss, the discrimination score for the masked image $\inputImg_n$ is computed by masking the elements of the discrimination map that correspond to patches containing invalid (masked) pixels, while computing the BN statistics in each layer only over features that are not affected by invalid input pixels (see SM for the importance of BN masking). Note that when computing the discrimination score for a fake image~$\fake$, there are no invalid pixels to ignore.
%\tamar{lets include an ablation of what happens without all these (we can put it in the supplementary)} \noa{added in SM examples when calculating BN with all pixels}.
 %for scales belong to $\Gcoarse$, and the mean of the \emph{masked} discrimination map at scales belong to $\Gfine$. (ii) 

The goal of the \emph{reconstruction loss} is to guarantee that there is a fixed input noise $\noiserec_n$, which is mapped to an image that equals $\inputImg_n$ at the valid regions. Therefore, we take this loss to be the MSE between $G_n(\noiserec_n)\odot(1-\tilde{\mask}_n)$ and $\inputImg_n\odot(1-\tilde{\mask}_n)$, where $\tilde\mask_n$ is a soft version of $\mask_n$, whose boundaries gradually transition from $0$ to $1$. %the gradual version of $\mask_n$, that is, we blur the boundaries of the binary map to create a gradual.
The reconstruction loss stabilizes training, and is also crucial for generating our completions at inference time (see Sec.~\ref{sec:inference} below). %At $\Gcoarse$ this is done by minimizing the MSE between $G_n(\noiserec_n)$ and $x_n$ (where $x_n$ is (optionally downsampled version of) the naive inpainting approximation). At $\Gfine$ this is done by minimizing the MSE between the valid area of the input image $\real_n\odot\tilde(1-\mask_n)$ and the corresponding valid area of the reconstruction image $G_n(\noiserec_n)\odot\tilde(1-\mask_n)$. Here $\tilde\mask_n$ is a gradual version of $\mask_n$, that is, we blur the boundaries of the binary map to create a gradual 
%\tamar{maybe add an ablation of w/ and w/o the gradual mask?}
%\tamar{maybe }
% In SinGAN, a noise \Noiserec  is fixed at the beginning of the training at the coarsest scale, and is mapped to the real image \Real  using a reconstruction loss as described in equation \eqref{eq:rec_loss_SinGAN}. In all others scales it is fixed to zeros.

%\begin{equation}
%  \begin{aligned}
%    \label{eq:rec_loss_SinGAN}
%        \fake^{rec}_{N} &= G_N({\noiserec}_N) \\
%        \fake^{rec}{_n} &=  G({\noiserec}_n,\fake_{n-1}\uparrow^{\alpha_n}), n = N-1,...,0 \\
%     loss &= MSE(\real_n ,  \fake^{rec}_{n}) \\
%     loss &= MSE(\real_n \odot {\tilde\mask}_n, \fake^{rec}{_n}  {\tilde\Mask}_n) \\
%  \end{aligned}
% \end{equation}
% \tamar{not sure we need these either. If we do keep these, the loss should be in a separate eq., including the adv. loss.}

Ideally, we would want all GANs in the pyramid to directly learn only from the valid patches in $\inputImg_n$ (those not containing masked pixels), as described above. However, when the masked region is large, there often do not exist sufficiently many valid patches for stably training a GAN, especially at the coarse scales. This is illustrated in Fig.~\ref{fig:patches_disc}, where only patches whose center pixel is outside the black region, can be used for training. We, therefore, treat the coarse scales differently from the fine ones.

\paragraph{Coarse scales}
Let $i$ be the finest scale in which less than 40\% of the patches are valid. In all scales belonging to $\{i,\ldots,N\}$, we do not perform masking in the discriminator. Instead, we use a naively inpainted version of the input as our real image, which we obtain using an internal technique. This naive inpainting solution is generated at scale $i$ and down-sampled to yield the real image at all coarser scales (see Fig.~\ref{fig:architecture}). Specifically, to obtain a naive completion, we adopt the deep image prior (DIP) method~\cite{ulyanov2018deep}, and modify it to ensure color consistency (see SM for the importance of the modification). Specifically, we train a light-weight U-Net model that maps a fixed input (random noise) into an inpainted image. This is done by minimizing two loss terms: (i) MSE  between the image $\inputImg_i$ and the U-Net's output over the valid pixels, and
(ii) MSE between each pixel in the inpainted region, and its nearest neighbor (in terms of RGB values) among the valid pixels in $\inputImg_i$. % \noa{this loss is only between pixels inside the masked area and $y_i$}
The latter term serves as an approximation to the KL divergence between the distributions of RGB values in the inpainted pixels and in the valid pixels~\cite{michaeli2014blind}. %The loss is described in \eqref{eq:skip}. \tamar{I'd omit this equation from the text. we can consider putting it in the SM.}
Finally, we stitch the naively inpainted region with the valid image regions. %between $\real_\scaletostartinpainting$ in the background and $\real'_\scaletostartinpainting$ in the missing area as shown in \eqref{eq:merge_inpaint}. 
Examples of naively inpainted solution and ablation of the method components %\noa{and the importance of the coarse scales} 
are provided in the SM. %Figure 

\subsection{Inference} \label{sec:inference}
Once the model is trained, our goal is to generate only the missing region. %, the pyramid of GANs is fixed and in order to generate a new fake image \Fake, we sample a noise $\noise_n$ in each scale (as in training) which is inserted into the generator. The output of the generator $\fake_n$ is upsampled to the size of $\real_{n-1}$ and goes into the generator with $\noise_{n-1}$. The output of the generator at the finest scale is our fake image $\fake_0$.
%However, in the task of inpainting we are not interested in sampling a whole new image but only the missing pixels in the image, 
Thus, instead of sampling full noise maps (as done during training), we construct noise maps containing $\noiserec_n$ at all elements affecting the valid regions, and new noise samples $\noise_n$ at the rest of the elements. Note that each element of the noise map affects a region of the size of the receptive field in the generated output. Therefore, the new noise samples are generated only within an eroded version of the mask (by half a receptive field), so that the noise map is constructed as $\noise^\text{test}_n = \noiserec_n \odot (1-\text{erode}\{\mask_n\}) + \noise_n\odot\text{erode}\{\mask_n\}$. This is illustrated in Fig.~\ref{fig:inference}. 
% See SM for ablation of controlling diversity of the results.% and generate a new noise sample only inside the mask, \ie the input noise map at each scale is equal to $\noise^\text{inference}_n = \noiserec_n \odot (1-\mask_n) + \noise_n\odot\mask_n$.% This is described in fig.~\ref{fig:inference_model}.
The final completion result is achieved by merging the generated inpainted region with the input image outside the mask. Since we want the fusion to be smooth, for this stage we use a soft version of the mask, obtained by convolving it with a Gaussian kernel with $\sigma=5$. 
% \tomer{not $\sigma=5$?}\noa{yes}. 
% The final inpainting solution is presented in \eqref{eq:inpaint}.
\input{figures/output_dis}

\input{figures/inference}

%% file: figures/architecture.tex
\begin{figure*}[t]
\centering
\includegraphics[width=1.85\columnwidth]{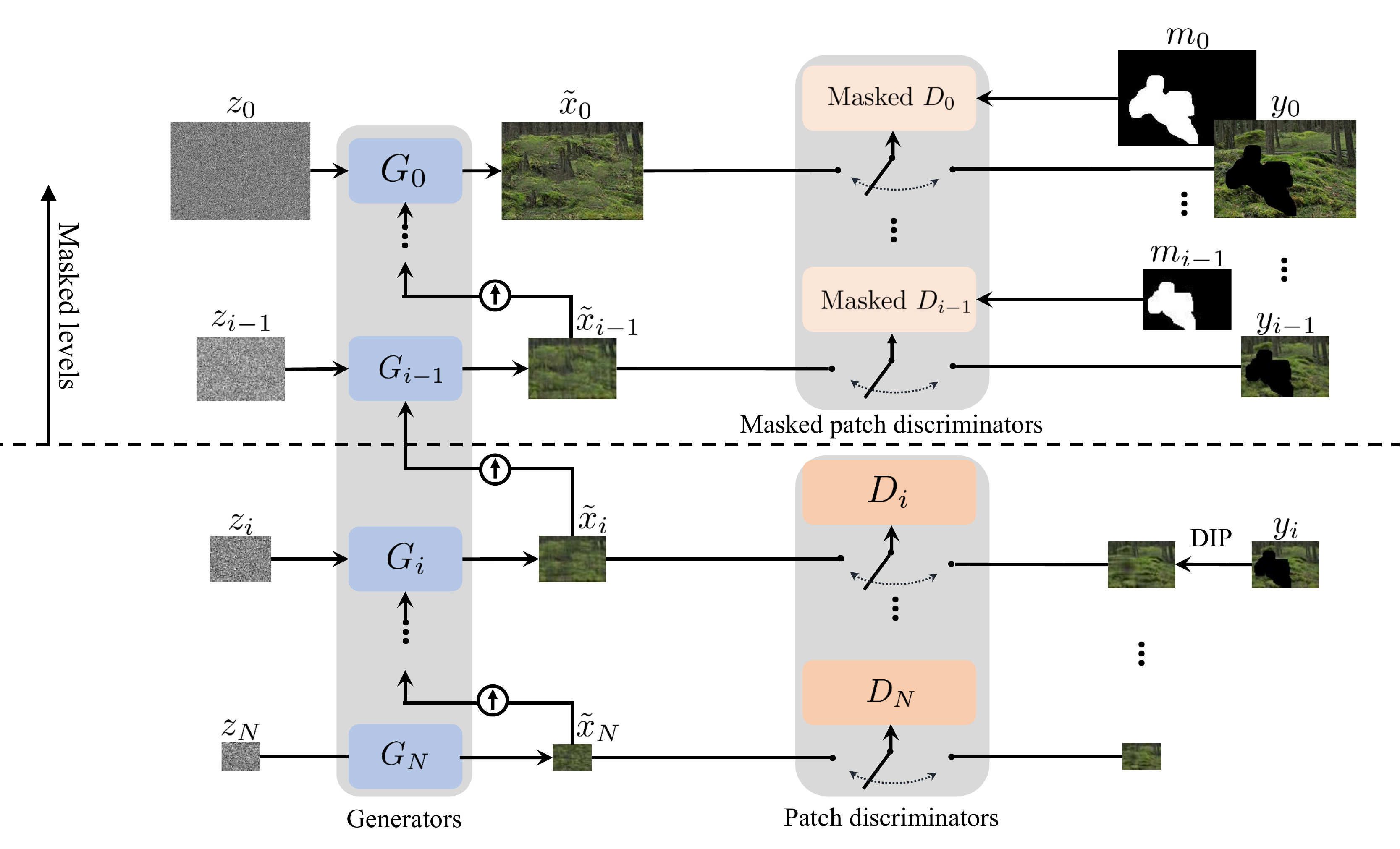}
	\caption{\label{fig:architecture}\textbf{Architecture.} Our model is composed of a pyramid of GANs, which we train sequentially from coarse to fine. Each generator consumes a noise map and the upsampled version of the image generated by the preceding scale, and it outputs a fake image. Each discriminator learns to distinguish generated images from the real one. At coarse scales, the real image presented to the discriminator is a naively inpainted version of the masked image. At fine scales, the real image is the masked image itself, and the discriminator ignores the missing region.
% 	\tamar{I think that we should consider describing the masking operation in a figure, either here somehow, or in a separate figure.}
}
\end{figure*}

%% file: figures/output_dis.tex
\begin{figure}[t]
\centering
\includegraphics[width=1\columnwidth]{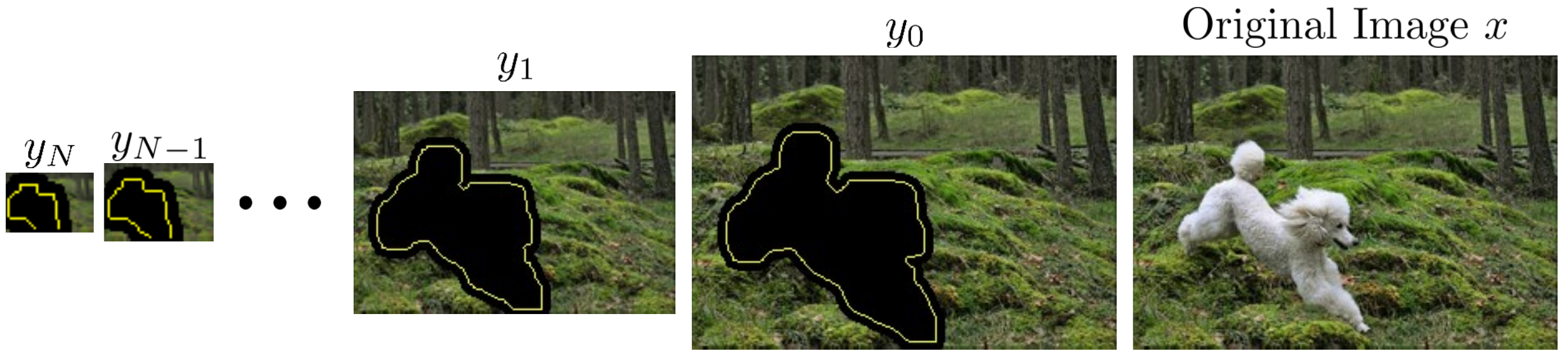}
	\caption{\label{fig:patches_disc}\textbf{Patches available for training in each scale.} We want the discriminator to classify only patches not containing missing pixels and thus dilate the mask (yellow) by half a receptive field. Pixels outside the black region are the centers of valid patches. At coarse scales, the number of valid patches is insufficient for training. Therefore, at the coarse scales, we perform naive inpainting rather than masking.
}
\end{figure}

%% file: figures/inference.tex
\begin{figure}[t]
\centering
    \begin{subfigure}[t]{1\columnwidth}
	    \vspace{-3mm}
		\centering
		\includegraphics[width=0.8\linewidth]{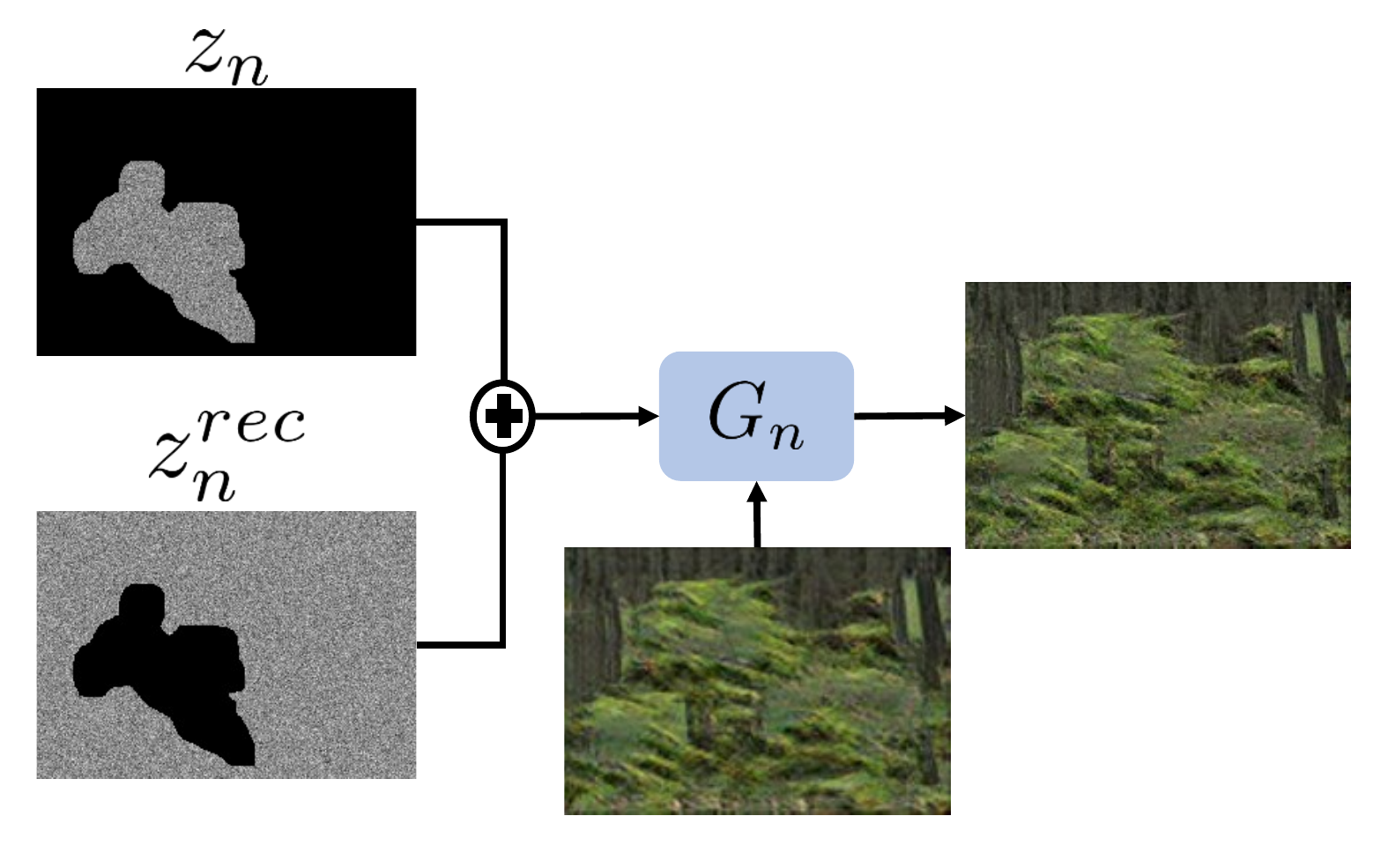}
		\caption{Inference}
	\end{subfigure}
	\begin{subfigure}[t]{1\columnwidth}
	    \vspace{-1mm}
		\centering
		\includegraphics[width=0.8\linewidth]{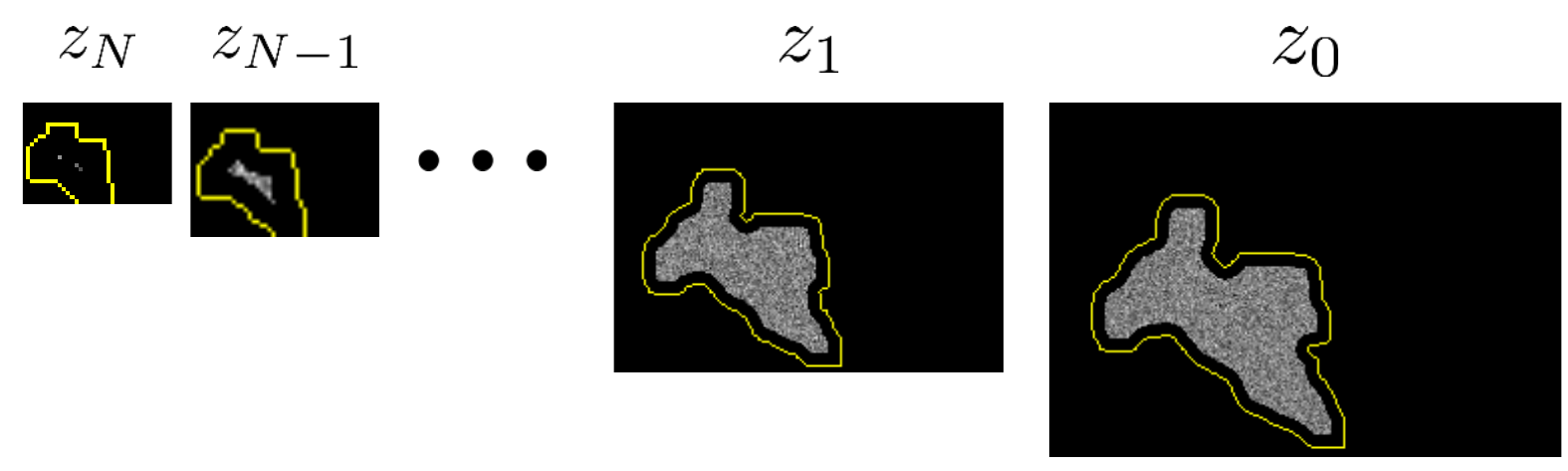}
		\caption{Sampled area at each scale}
	\end{subfigure}
% 	\begin{subfigure}[t]{1\textwidth}
% 	   % \vspace{-.3mm}
% % 		\centering
% 		\includegraphics[width=0.5\linewidth]{./figures/inference.pdf}
% 		\caption{Inference}
% 	\end{subfigure}
%     \begin{subfigure}[t]{1\textwidth}
% 	    \vspace{-.3mm}
% % 		\centering
% 		\includegraphics[width=0.5\linewidth]{./figures/methods_inference.pdf}
% 		\caption{Sampled area at each scale}
% 	\end{subfigure}

% 	 This way, since we sample a smaller area the diversity will be smaller, but we will keep fixed the knowing pixels in the image. However, the erosion of the masks can vary from half a RF (as in the first row) to zero pixels (as in  the last row). When the erosion is smaller, the variability of the results is higher. The last column represent the affected area in the final image at each approach and describes the variance of each pixel over several samples. The affected area is calculated by taking k \todo{right the number itself instead of k} samples and calculating the variance of each pixel in the image.
% 	(b) 
% 	\tamar{I still think (b) should be much larger and also all the fonts. Also in (b) it is hard to see any noise. In (a) why some of the masks have halos? I think (a) should come after (b) (change there order, as appeared in the text.)}}
\caption{\label{fig:inference}
\textbf{Inference.} (a) At inference time, we sample $z_{n}$ inside the mask and use $z^{\text{rec}}_{n}$ outside the mask. The combined noise map is injected to the generator together with the up-sampled fake image from the previous scale. (b) The noise $z_{n}$ is sampled only within regions that do not affect the valid part of the image. This requires eroding the mask (yellow) by half a receptive field.}

\end{figure}

%% file: sections/4_experiments.tex
\section{Experiments}

%\tamar{I added subsections and paragraph headlines. I think it will ease the reading for the tired reviewer. You can remove of course (or regard this as adversarial attack =) ).}

We now evaluate the performance of our method, qualitatively and quantitatively, and compare it to existing baselines. We focus on two scenarios: \emph{object removal} and \emph{generic inpainting}. In the former, which is commonly encountered in real applications, the goal is to edit an image so as to cut out a whole semantic object. In the latter, which is the focus of many papers, the goal is to treat arbitrary masks (not necessarily covering a whole object). %To simulate this task, we use the Part-Imagenet~\cite{he2021partimagenet} dataset, which contains images of single objects, along with their corresponding segmentation maps. We use the segmentation map as our mask. 
%\tamar{I moved the previous sentence to the next subsection. Also I suggest to add here:} "
As we show, our approach performs at least comparable to the baselines, especially in the task of object removal. This is despite the fact that it has access only the masked image. %In addition we evaluate our method for the task of inpainting arbitrary masks (not necessarily covering an object). Although our method is not targeted for this task, its performance is usually rather close to external methods."

\subsection{Object removal}
The common practice in evaluating generic inpainting methods is to randomly mask regions within images (say from Places~\cite{zhou2014learning} or ImageNet~\cite{deng2009imagenet}) and use the non-masked images as ground-truth. %\tamar {I'd remove the beginning of the sentence "It is important to note that" and just start from "The common...". Feels a bit too apologizing currently.}
However, this does not simulate well the object removal task, as it results in cutting out of parts of semantic objects.
To simulate the object removal task, here we use the Part-Imagenet~\cite{he2021partimagenet} dataset, which contains images of single objects, along with their corresponding segmentation maps. We use the segmentation map as our mask. 
%The common practice in evaluating generic inpainting methods is to randomly mask regions within images (say from Places~\cite{zhou2014learning} or ImageNet~\cite{deng2009imagenet}) and use the non-masked images as ground-truth. %\tamar {I'd remove the beginning of the sentence "It is important to note that" and just start from "The common...". Feels a bit too apologizing currently.}
%However, this does not simulate well the object removal task, as it results in cutting out of parts of semantic objects. The problem is that in object removal, there exist no ground-truth data to compare to (\ie we have no access to an image of the background behind the object), and therefore evaluation is more challenging. \tamar{maybe remove the last part "and therefore evaluation is more challenging". I think it is enough to say that there is no gt image.}

% \textbf{Qualitative evaluation}
% Qualitative results are shown in fig.~\ref{fig:results_all_methods}, fig.~\ref{fig:partImagenet_nondiverse} and fig.~\ref{fig:partImagenet_diverse}. 
\myparagraph{%We begin with
Qualitative comparisons} Figure~\ref{fig:partImagenet_nondiverse} shows completion results generated by our method as well as other inpainting techniques: the non-diverse external methods %contextual attention (
CA~\cite{yu2018generative} and LAMA~\cite{suvorov2022resolution}, the diverse external methods DSI~\cite{peng2021generating}, PIC~\cite{zheng2019pluralistic}, ICT~\cite{wan2021high} and CoMod-GAN~\cite{zhao2021large}, %which proposes two-stage model. For both methods when evaluating on images from places2 dataset, we used models trained on Places2 with a random mask for the centered square missing part with size of $85 \times 85$ and a model trained specifically only with centered square missing part with size of $128 \times 128$. When evaluating their performance on Part-Imagenet we used the models trained on ImageNet dataset with random masks. 
% (ii) Non-Diverse external methods: %contextual attention (
% CA~\cite{yu2018generative}, %which %was trained on an external training set and 
% relays on . %layers in a refinement network. This method suggests a single inpainting solution. 
and the non-diverse internal methods %to methods which do not rely on a training set, but uses only the image to be inpainted itself. Such methods are
%Telea~\cite{telea2004image}, %which uses a fast marching method, 
%naiver-stocks (NS) based method
%NS~\cite{bertalmio2001navier} , %from fluid dynamics for propagating information, 
%Frequency Selective Reconstruction 
%\todo{the full name} 
% FSR~\cite{seiler2015resampling}, %which assumes that small areas in natural images can be sparsely modeled in the Fourier domain,
Shift-Map~\cite{pritch2009shift}, Patch-Match~\cite{barnes2009patchmatch} and DIP~\cite{ulyanov2018deep}. % which represents the image and maps pixels inside the mask from different locations in the image, and Deep Image Prior 
% DIP~\cite{ulyanov2018deep}. 
We use ImageNet models where available (PIC, DSI, and ICT).
% For each diverse-external method we show results with a model trained on Places and with a model trained on ImageNet. 
See SM for comparisons with other models and methods. % which overfits a neural network to the valid area of the image to be inpainted. Note that all these methods do not offer diverse solutions, but rather a single one.
% \todo{update figure~\ref{fig:results_all_methods}, and update the text accordingly.}%inpainting result for all the methods we are comparing to. One can see that 
% As can be seen, our method is better than most internal methods in term of visual quality (except for Shift-Map which is comparable), as these usually produce irregularity at the completion borders. In addition our performances are comparable to the external methods. Please see many additional qualitative comparison in the Supplementary Material (SM).
%\input{figures/inference}
% \input{figures/partImagenet_nondiverse}
As can be seen, 
%This figure demonstrates the advantage of internal methods over external one in the context of object removal. E
externally trained methods %tend to complete the image in a way that matches the distribution of their training set (\ie ImageNet~\cite{deng2009imagenet}) %, Places~\cite{zhou2014learning}
%, and %Therefore they 
sometimes generate an object in the missing region, which is undesired in an object-removal setting. %This may %which %This 
%result in unrealistic completions and artifacts.
Patch-Match and DIP often generate blurry completions. Shift-map fills the missing pixels by copying information from other parts of the image and therefore leads to more realistic results, however it does not offer diverse solutions. Our method generates plausible completions, and suggests diverse solutions (Fig.~\ref{fig:partImagenet_diverse}).
%In fig.~\ref{fig:partImagenet_diverse} we illustrate the capability of our method to produce diverse solutions comparing to other diverse methods. 
Note that our method is trained on very limited information (only the non-masked pixels within the image). Yet, it manages to produce diverse completions that are at least comparable in visual quality to diverse external methods.
\input{figures/partImagenet_nondiverse_latex}

\input{figures/partImagenet_comparison}

%\textbf{Human perceptual study}
\myparagraph{Human perceptual study}
%Next, 
We quantify the realism of our results using a user study performed through the Amazon Mechanical Turk platform. We randomly chose 50 images from the Part-Imagenet dataset and conducted 13 surveys, each comparing our method with a different baseline. All images participating in the studies appear in the SM. %other competitive methods from each category: (i) Non-Diverse External: CA (ii) Diverse External: PIC and DSI (with two variations of each, trained on ImageNet and Places2) (iii) Non-Diverse Internal: Shift-Map and DIP. 
In each study, we included 50 questions: 5 tutorial trials (with a clear correct answer and feedback to the users) and 45 test questions. Each question displayed the original image, the masked image, and two possible completions: one of our method and one of the competing method. Users were asked to choose which completion they prefer. In total, 50 workers participated in each study. The results are shown in Fig.~\ref{fig:user_study_partImagenet}.  %\todo{add table}. 
For the DIP algorithm, we used both the original implementation, which runs for a fixed number of iterations, and an optimized variant that selects the result with the lowest loss along the algorithm's iterations. 
% \noa{add new experiments}
As can be seen, users tend to prefer our completions over 
other internal methods (ranking IDC as better than DIP and Patch-Match and comparable to Shift-Map). IDC is also ranked at least on par with most external methods, except for the recent LAMA and CoMod-GAN methods.
%those of %more than the completion generated by 
%PIC (both variants) and DIP (both variants). When compared against all other methods, users rank our completions as slightly inferior when comparing to LAMA and CoMod-GAN and comparable with the rest.
This is while our method has the advantage of being applicable also to other domains.%, and it offers a diverse solution (in contrast to CA, LAMA, shift-map and DIP).

\input{figures/boxplot}
\input{tables/partImagenet_quant}

%\textbf{Quantitative evaluation}
\myparagraph{Quantitative evaluation}
We complement the user study with further quantitative evaluation. Since no ground truth is available in the object removal task (\ie we do not have images of the background behid the object), measures like PSNR and FID, which require reference images, cannot be used. We therefore use two common no-reference metrics: the naturalness image quality evaluator (NIQE)~\cite{niqe} and the neural image assessment (NIMA) measure~\cite{talebi2018nima}. Results are presented in Tab.~\ref{table:partImagenet_quant}. As can be seen, in terms of NIQE (lower is better) our method is better than other internal methods and slightly inferior to external methods, though this does not always align with users' preferences as can be seen in Fig~\ref{fig:user_study_partImagenet}. In terms of NIMA (higher is better) our method is ranked higher than DIP, PIC, CoMod-GAN, and DSI (Places model) and on par with all other baselines.

\myparagraph{Diversity}
We also quantify the semantic diversity among the different completions of each method, using the learned perceptual image patch similarity (LPIPS) measure~\cite{zhang2018unreasonable}. For each input image, we generate 20 pairs of diverse completions, and calculate the average LPIPS score between all pairs over all images. The results are reported in Tab.~\ref{table:partImagenet_quant}. As can be seen, our semantic diversity is a bit lower than most external methods. That is, the different completions we generate are not interpreted by a classification network as having very different semantic meanings. However, note that higher semantic diversity %We believe this steams from the fact that % Note that 
%higher diversity sometimes indicates %not always better since high diversity can result in 
 %sometimes results that are inadequate %not adequate 
is often indicative of insertion of new objects into the missing region, which is not desired in an object completion task (see \eg Fig.~\ref{fig:partImagenet_nondiverse} and Fig.~\ref{fig:partImagenet_diverse}). In any case, it is possible to control the diversity in IDC either by adjusting the noise levels (which has a small effect), or by adjusting the kernel used to erode the mask in each scale. We elaborate on these mechanisms in the SM. %\tamar{not sure about this...}. Moreover, lower diversity is acceptable in our internal method, since the completion of the image is limited only for statistics learned within the image. \tamar{too apologizing?} Nevertheless, we offer two additional operation modes with higher diversity at inference (\ie no additional training is required), so the user can chose the desired diversity level.

\input{figures/results_all_methods}

\input{tables/realism_and_diversity}
\subsection{Generic inpainting}
Although our main goal is object removal, we now provide comparisons to baselines on the task of generic inpainting (\ie inpainting with arbitrary masks). In this setting, the mask may occlude parts of semantic objects, which internal methods cannot complete (see SM). However, as we now show, the performance of our method is usually rather close to externally trained models, even on such tasks. %which were trained particularly for this task. 
%In order to compare ourselves to the specific task that external methods were trained for, 
Here we use the Places validation dataset~\cite{zhou2014learning}, with a missing region of a centralized square of size $85 \times 85$ or $128\times128$. In this setting, since we have ground truth images, we calculate The Fr\'echet Inception Distance  (FID)~\cite{heusel2017gans} in addition to NIQE and NIMA. % which were used also in Part-ImageNet comparison.
% structural similariy (SSIM) and peak signal to noise ratio (PSNR). 
Results are presented in Tab.~\ref{table:realism_diversity}. 
As can be seen, for both mask sizes, our method achieves a better FID score than DIP and is comparable to Shift-Map and Patch-Match. Note, however, that Shift-map often generates recognizable transitions at boundaries of the inpainted region, which FID does not capture and that the completions of Patch-Match are often blurrier (see Fig.~\ref{fig:results_all_methods}). 
%\noa{remove ref if we remove the fig}).
Our results are only slightly inferior to external methods, despite the fact that they were trained on thousands of images from this specific dataset.

%% file: figures/partImagenet_nondiverse_latex.tex
\begin{figure*}[t]
	\centering
	\captionsetup[subfigure]{labelformat=empty,justification=centering,aboveskip=1pt,belowskip=0pt,font=scriptsize}
	%%%%%%%%%%%%%%%%%
	\begin{subfigure}[t]{0.13\textwidth}
	    \vspace{-.3mm}
		\centering
		\caption{Original}
		\includegraphics[width=1\linewidth]{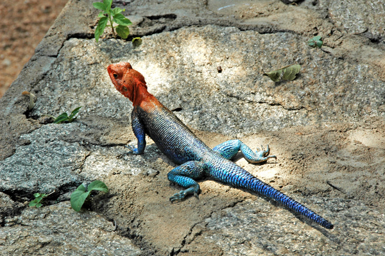}
	\end{subfigure}
	\begin{subfigure}[t]{0.13\textwidth}
	    \vspace{-.3mm}
		\centering
		\caption{Input}
		\includegraphics[width=1\linewidth]{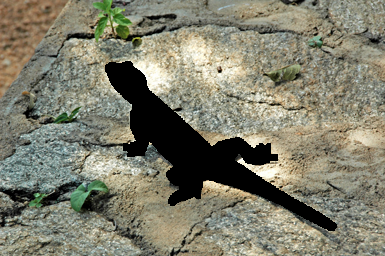}
	\end{subfigure}
	\begin{subfigure}[t]{0.13\textwidth}
	    \vspace{-.3mm}
		\centering
		\caption{CA}
		\includegraphics[width=1\linewidth]{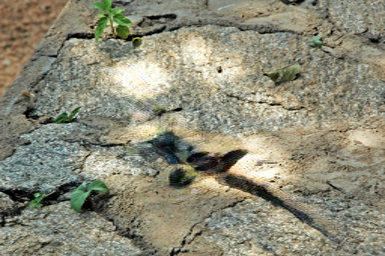}
	\end{subfigure}
		\begin{subfigure}[t]{0.13\textwidth}
	    \vspace{-.3mm}
		\centering
		\caption{LAMA}
		\includegraphics[width=1\linewidth]{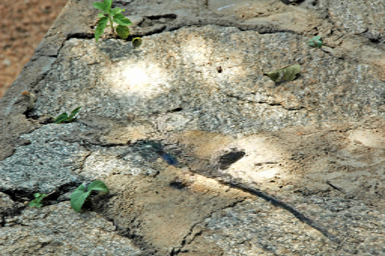}
	\end{subfigure}
		\begin{subfigure}[t]{0.13\textwidth}
	    \vspace{-.3mm}
		\centering
		\caption{DSI}
		\includegraphics[width=1\linewidth]{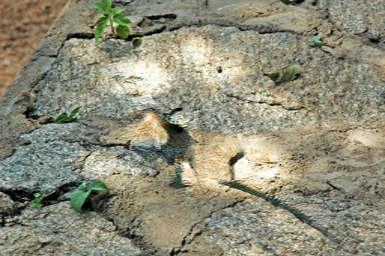}
	\end{subfigure}
		\begin{subfigure}[t]{0.13\textwidth}
	    \vspace{-.3mm}
		\centering
		\caption{PIC}
		\includegraphics[width=1\linewidth]{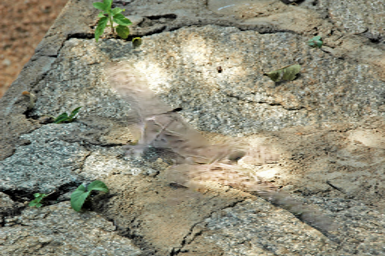}
	\end{subfigure}

	\begin{subfigure}[t]{0.13\textwidth}
	    \vspace{-.3mm}
		\centering
		\caption{ICT}
		\includegraphics[width=1\linewidth]{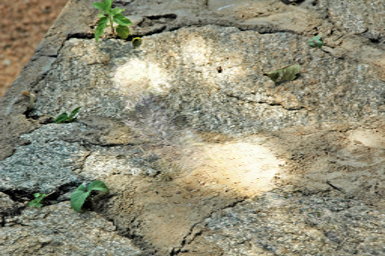}
	\end{subfigure}
	\begin{subfigure}[t]{0.13\textwidth}
	    \vspace{-.3mm}
		\centering
		\caption{CoMod-GAN}
		\includegraphics[width=1\linewidth]{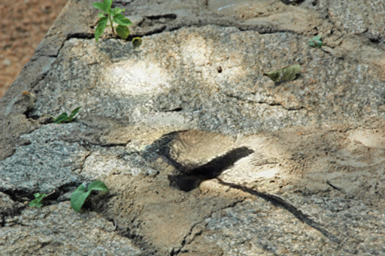}
	\end{subfigure}
	\begin{subfigure}[t]{0.13\textwidth}
	    \vspace{-.3mm}
		\centering
		\caption{Shift-Map}
		\includegraphics[width=1\linewidth]{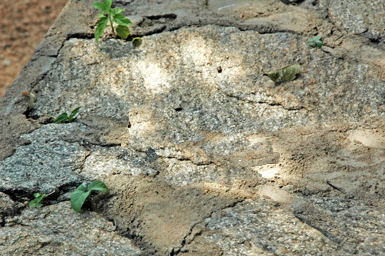}
	\end{subfigure}
	\begin{subfigure}[t]{0.13\textwidth}
	    \vspace{-.3mm}
		\centering
		\caption{Patch-Match}
		\includegraphics[width=1\linewidth]{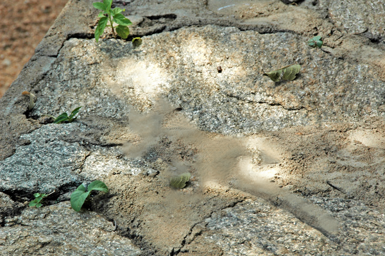}
	\end{subfigure}
		\begin{subfigure}[t]{0.13\textwidth}
	    \vspace{-.3mm}
		\centering
		\caption{DIP}
		\includegraphics[width=1\linewidth]{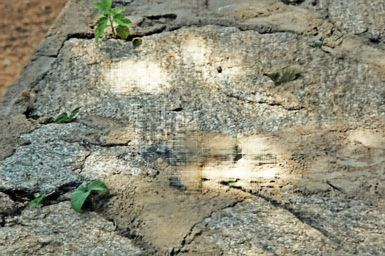}
	\end{subfigure}
		\begin{subfigure}[t]{0.13\textwidth}
	    \vspace{-.3mm}
		\centering
		\caption{IDC (ours)}
		\includegraphics[width=1\linewidth]{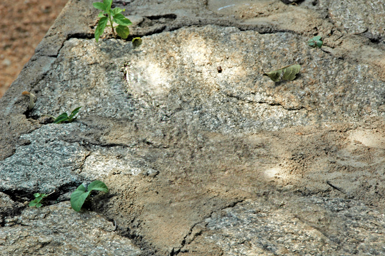}
	\end{subfigure}

 	\begin{subfigure}[t]{0.13\textwidth}
 	    \vspace{-.3mm}
 		\centering
 		\caption{Original}
 		\includegraphics[width=1\linewidth]{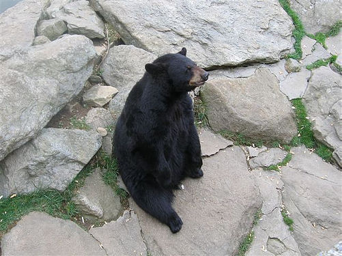}
 	\end{subfigure}
 	\begin{subfigure}[t]{0.13\textwidth}
 	    \vspace{-.3mm}
 		\centering
 		\caption{Input}
 		\includegraphics[width=1\linewidth]{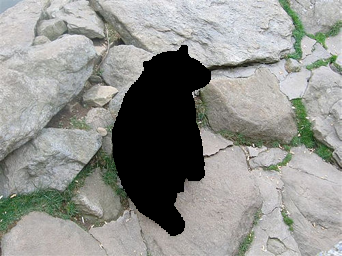}
 	\end{subfigure}
 	\begin{subfigure}[t]{0.13\textwidth}
 	    \vspace{-.3mm}
 		\centering
 		\caption{CA}
 		\includegraphics[width=1\linewidth]{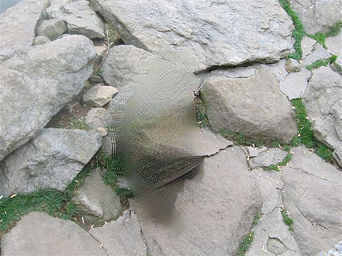}
	\end{subfigure}
		\begin{subfigure}[t]{0.13\textwidth}
	    \vspace{-.3mm}
		\centering
		\caption{LAMA}
		\includegraphics[width=1\linewidth]{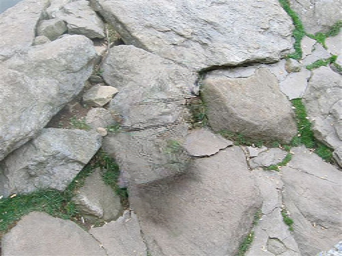}
	\end{subfigure}
		\begin{subfigure}[t]{0.13\textwidth}
	    \vspace{-.3mm}
		\centering
		\caption{DSI}
		\includegraphics[width=1\linewidth]{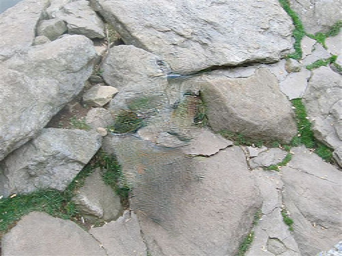}
	\end{subfigure}
		\begin{subfigure}[t]{0.13\textwidth}
	    \vspace{-.3mm}
		\centering
		\caption{PIC}
		\includegraphics[width=1\linewidth]{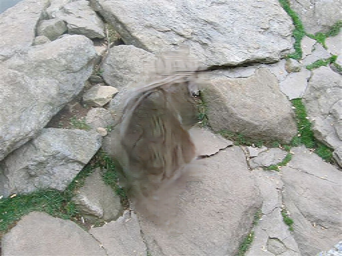}
	\end{subfigure}

			\begin{subfigure}[t]{0.13\textwidth}
	    \vspace{-.3mm}
		\centering
		\caption{ICT}
		\includegraphics[width=1\linewidth]{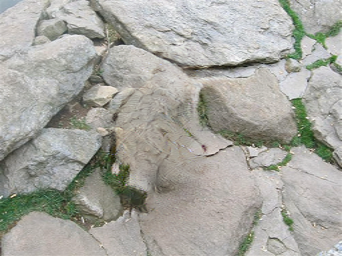}
	\end{subfigure}
	\begin{subfigure}[t]{0.13\textwidth}
	    \vspace{-.3mm}
		\centering
		\caption{CoMod-GAN}
		\includegraphics[width=1\linewidth]{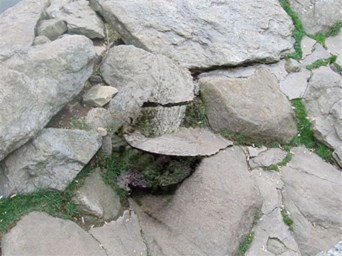}
	\end{subfigure}
	\begin{subfigure}[t]{0.13\textwidth}
	    \vspace{-.3mm}
		\centering
		\caption{Shift-Map}
		\includegraphics[width=1\linewidth]{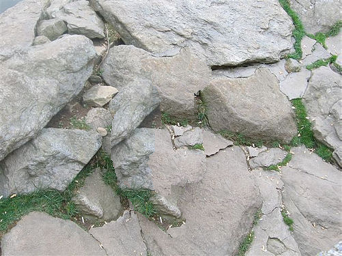}
	\end{subfigure}
	\begin{subfigure}[t]{0.13\textwidth}
	    \vspace{-.3mm}
		\centering
		\caption{Patch-Match}
		\includegraphics[width=1\linewidth]{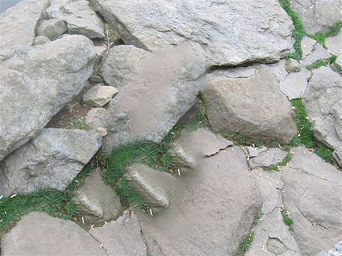}
	\end{subfigure}
		\begin{subfigure}[t]{0.13\textwidth}
	    \vspace{-.3mm}
		\centering
		\caption{DIP}
		\includegraphics[width=1\linewidth]{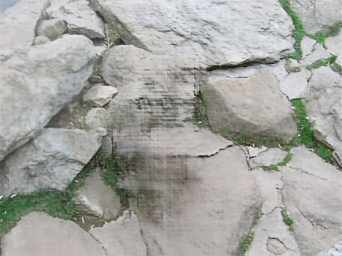}
	\end{subfigure}
		\begin{subfigure}[t]{0.13\textwidth}
	    \vspace{-.3mm}
		\centering
		\caption{IDC (ours)}
		\includegraphics[width=1\linewidth]{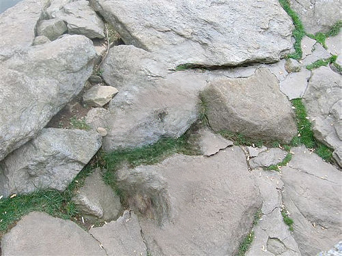}
	\end{subfigure}

	%%%%%%%%%%%%%%%
	%\setcounter{figure}{0}
	\caption{\label{fig:partImagenet_nondiverse}\textbf{Qualitative comparison.} Our method is at least comparable to baselines in terms of visual quality, while being the only diverse method that is applicable to arbitrary domains. %For DSI, PIC and ICT, we show ImageNet models and for the rest models trained on Places.
	See SM for completions with other models.
	\vspace{-0.5mm}
	}
\end{figure*}{}

%% file: figures/partImagenet_comparison.tex
\begin{figure*}[t]
	\centering
	\includegraphics[width=1.72\columnwidth]{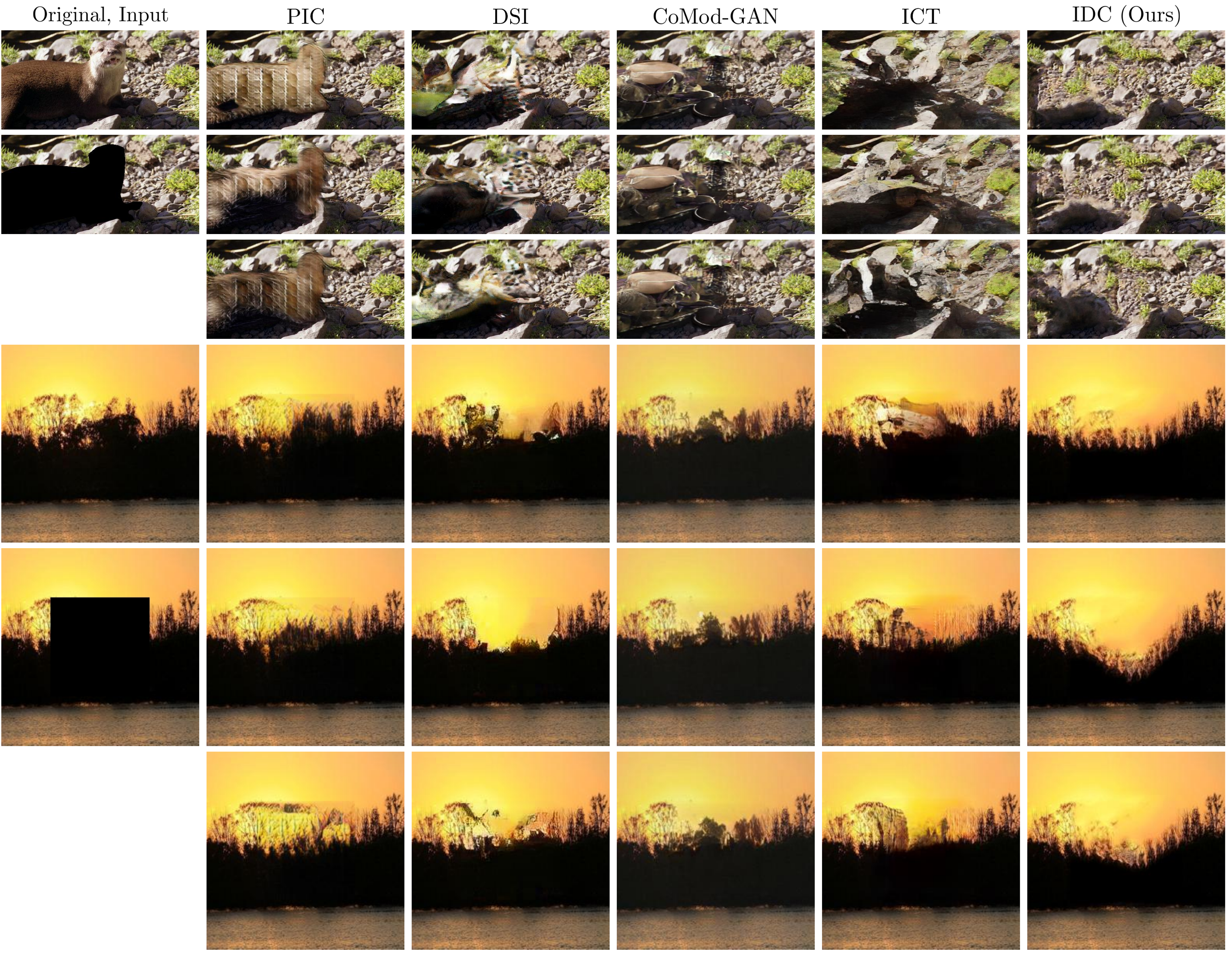}
		\caption{\label{fig:partImagenet_diverse}\textbf{Internal vs.~external diverse completion.} %We compare internal and external diverse completion. %For external methods, first image is taken from Part-Imagenet and uses a model trained on ImageNet. The rest are images from Places and uses a model trained on Places. One can see that e
		External methods complete the missing region according to the distribution of their training set (we use ImageNet models in the 1st example when available, and Places models in the 2nd). This may result in undesirable effects in the task of object removal.} %so as to match the statistics of their training set. 
		%Our method is confined to any specific training set. Therefore we are able to offer many plausible distinct solutions regardless of the image type.}

\end{figure*}

%% file: figures/boxplot.tex
\begin{figure*}[t]
\includegraphics[width=1.8\columnwidth]{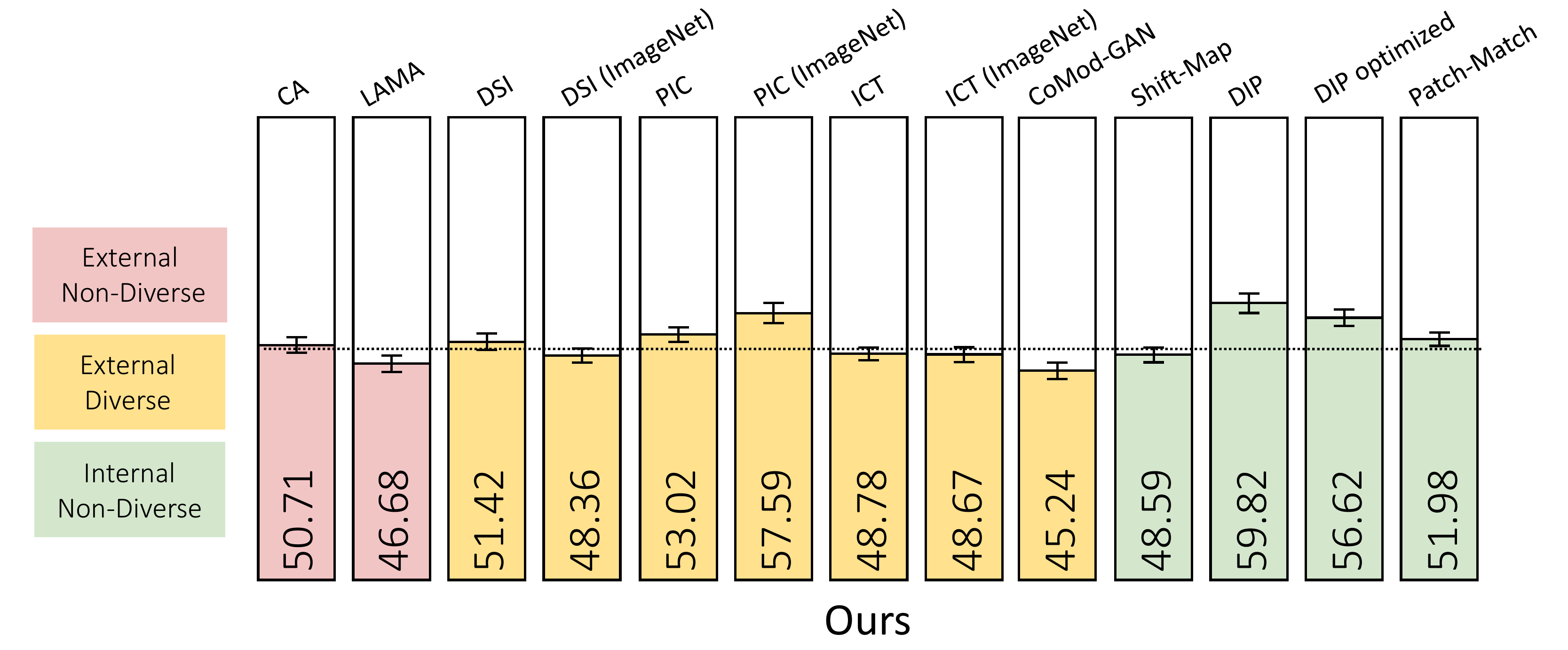}
\centering
    %\vspace{-0.7cm}
	\caption[Human perceptual studies.]{\label{fig:user_study_partImagenet}
	\textbf{Human perceptual studies.} We report user preferences for our completions over competing baselines %in two alternative . in completion task for 
        on Part-Imagenet (bootstrap is used for calculating standard deviation). %The results show that our completion win PIC and DIP methods and are comparable to the other methods, even external models. 
        We compare to external models trained on Places dataset and, when available, also to models trained on ImageNet. As can be seen, our method is ranked by users as comparable 
        % (black) 
        or better 
        % (green) 
        than most baselines, while it is the only diverse technique that is applicable to arbitrary domains (being internally trained).}
\vspace{-0.35cm}
\end{figure*}

%% file: tables/partImagenet_quant.tex
       \begin{table*}[h]
            \begin{center}
            \smallskip\noindent
            \resizebox{0.8\linewidth}{!}{
                \begin{tabular}{@{}clclcccccc@{}} 
                \toprule
                & & & & & NIQE$\downarrow$ & & NIMA$\uparrow$ & &  LPIPS $\cdot10^{-3}$\\
                & & & & & & & & & (Diversity)\\
                \midrule
                & Non-Diverse & & CA (Places) & & 5.19 & & 4.81 & & 0\\
                & & & LAMA (Places) & & 5.18 & & 4.77 & & 0\\
                \cmidrule{2-10}
                External &  & & DSI (Places\textbf{/}ImageNet) & & 5.17\,\textbf{/}\,5.24 & & 4.70\,\textbf{/}\,4.83 & & $48$ \textbf{/} $46$\\
                 & Diverse & & PIC (Places\textbf{/}ImageNet) & & 5.15\,\textbf{/}\,4.82 & & 4.74\,\textbf{/}\,4.76 & & $23$\,\textbf{/}\,$22$\\
                & & & ICT (Places\textbf{/}ImageNet) & & 5.10\,\textbf{/}\,5.19 & & 4.78\,\textbf{/}\,4.77 & & $55$\,\textbf{/}\,$62$\\
                & & & CoMod-GAN (Places)  & & 4.91 & & 4.73 & &  $13$\\
                \midrule
                 & Non-
                 & & Shift-Map & & 5.53 & & 4.81 & & 0\\
                 & Diverse & & DIP (original\textbf{/}optimized) & &  5.54\,\textbf{/}\,5.25 & &  4.33\,\textbf{/}\,4.44 & & 0\,\textbf{/}\,0\\
                Internal & & & Patch-Match & & 5.32 & & 4.82 & & 0\\ %$20$
                \cmidrule{2-10}
                & Diverse & & IDC (Ours) & & 5.27 & & 4.79 & & $15$\\
                \bottomrule
            
            \end{tabular}}
            
            % \begin{tabular}{@{}clclcccccc@{}} 
            %     \toprule
            %     & & & & & NIQE$\downarrow$ & & NIMA$\uparrow$ & &  LPIPS\\
            %     & & & & & & & & & (Diversity)\\
            %     \midrule
            %     \multirow{10}{4em}{External} & \multirow{2}{4em}{Non-Diverse} & & CA & & 5.19 & & 4.81 & & 0\\
            %     & & & LAMA & & 5.18 & & 4.77 & & 0\\
            %     \cmidrule{2-10}
            %     & \multirow{7}{4em}{Diverse} & & DSI (ImageNet) & & 5.24 & & 4.83 & & $46$\\
            %     & & & DSI (Places)  & & 5.17 & & 4.70 & & $48$\\
            %     & & & PIC (ImageNet) & & 4.82 & & 4.76 & & $22$\\
            %     & & & PIC (Places) & & 5.15 & & 4.74 & &  $23$\\
            %     & & & ICT (ImageNet) & & 5.19 & & 4.77 & & $62$\\
            %     & & & ICT (Places) & & 5.10 & & 4.78 & &  $55$\\
            %     & & & CoMod-GAN & & 4.91 & & 4.73 & &  $13$\\
            %     \midrule
            %     \multirow{3}{4em}{Internal} & \multirow{2}{4em}{Non-Diverse}
            %      & & Shift-map & & 5.53 & & 4.81 & & 0\\
            %     & & & DIP (optimized) & & 5.54 (5.25) & & 4.33 (4.44) & & 0\\
            %     \cmidrule{2-10}
            %     & \multirow{2}{4em}{Diverse} & & Patch-Match & & 5.32 & & 4.82 & & $20$\\
            %     & & & IDC (Ours) & & 5.27 & & 4.79 & & $15$\\
            %     \bottomrule
            
            % \end{tabular}}
        \end{center}
        \caption{\textbf{Quantitative evaluation.} We quantify visual quality and semantic diversity on the Part-Imagenet dataset. In external methods, we refer to Places models, and when available, also to ImageNet models.
        In terms of NIQE (lower is better), our method is ranked as better than all internal methods and slightly inferior to external methods (though this does not always align with users' preferences; see Fig.\ref{fig:user_study_partImagenet}). In  terms of NIMA (higher is better), our method is ranked as better than DIP, PIC, CoMod-GAN and DSI (Places model) and comparable to all other methods. The semantic diversity achieved by our method is similar to that of CoMod-GAN and somewhat lower than other external diverse inpainting approaches.
        }
        \label{table:partImagenet_quant}
        \end{table*}

%% file: figures/results_all_methods.tex
\begin{figure*}[t]
	\centering
	\captionsetup[subfigure]{labelformat=empty,justification=centering,aboveskip=1pt,belowskip=0pt,font=scriptsize}
	%%%%%%%%%%%%%%%%%
% 	\begin{subfigure}[t]{0.13\textwidth}
% 	    \vspace{-.3mm}
% 		\centering
% 		\caption{Original}
% 		\includegraphics[width=1\linewidth]{figures/comparsion_places_val_85/26.png}
% 	\end{subfigure}
	\begin{subfigure}[t]{0.13\textwidth}
	    \vspace{-.3mm}
		\centering
		\caption{Input}
		\includegraphics[width=1\linewidth]{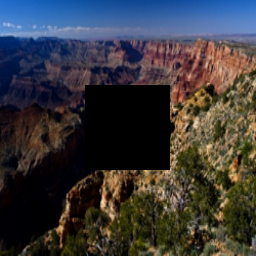}
	\end{subfigure}
	\begin{subfigure}[t]{0.13\textwidth}
	    \vspace{-.3mm}
		\centering
		\caption{CA}
		\includegraphics[width=1\linewidth]{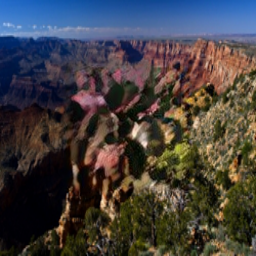}
	\end{subfigure}
		\begin{subfigure}[t]{0.13\textwidth}
	    \vspace{-.3mm}
		\centering
		\caption{LAMA}
		\includegraphics[width=1\linewidth]{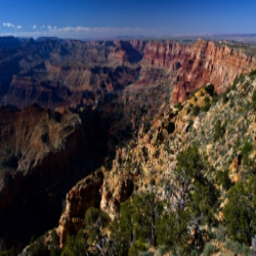}
	\end{subfigure}
		\begin{subfigure}[t]{0.13\textwidth}
	    \vspace{-.3mm}
		\centering
		\caption{DSI}
		\includegraphics[width=1\linewidth]{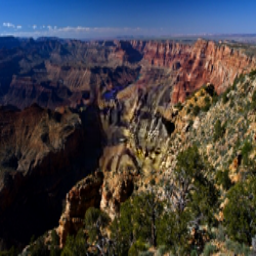}
	\end{subfigure}
		\begin{subfigure}[t]{0.13\textwidth}
	    \vspace{-.3mm}
		\centering
		\caption{PIC}
		\includegraphics[width=1\linewidth]{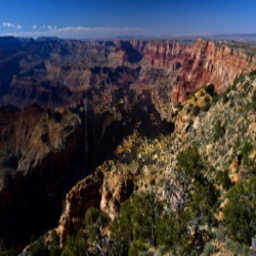}
	\end{subfigure}
			\begin{subfigure}[t]{0.13\textwidth}
	    \vspace{-.3mm}
		\centering
		\caption{ICT}
		\includegraphics[width=1\linewidth]{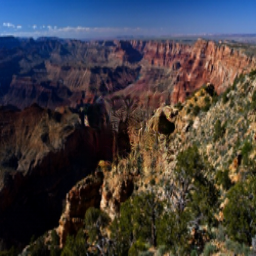}
	\end{subfigure}

	\begin{subfigure}[t]{0.13\textwidth}
	    \vspace{-.3mm}
		\centering
		\caption{CoMod-GAN}
		\includegraphics[width=1\linewidth]{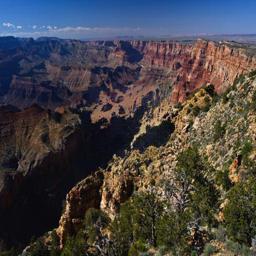}
	\end{subfigure}
	\begin{subfigure}[t]{0.13\textwidth}
	    \vspace{-.3mm}
		\centering
		\caption{Shift-Map}
		\includegraphics[width=1\linewidth]{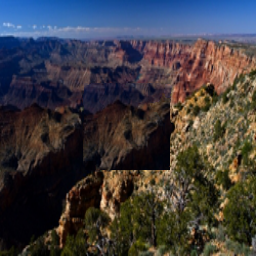}
	\end{subfigure}
	\begin{subfigure}[t]{0.13\textwidth}
	    \vspace{-.3mm}
		\centering
		\caption{DIP}
		\includegraphics[width=1\linewidth]{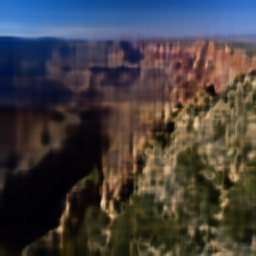}
	\end{subfigure}
		\begin{subfigure}[t]{0.13\textwidth}
	    \vspace{-.3mm}
		\centering
		\caption{DIP optimized}
		\includegraphics[width=1\linewidth]{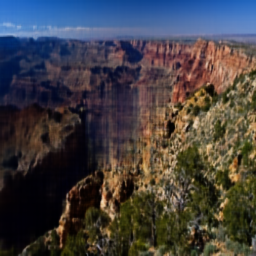}
	\end{subfigure}
		\begin{subfigure}[t]{0.13\textwidth}
	    \vspace{-.3mm}
		\centering
		\caption{Patch-Match}
		\includegraphics[width=1\linewidth]{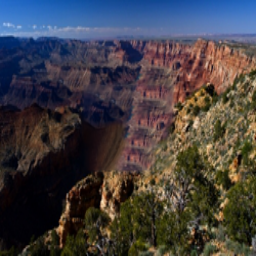}
	\end{subfigure}
	\begin{subfigure}[t]{0.13\textwidth}
	    \vspace{-.3mm}
		\centering
		\caption{IDC (Ours)}
		\includegraphics[width=1\linewidth]{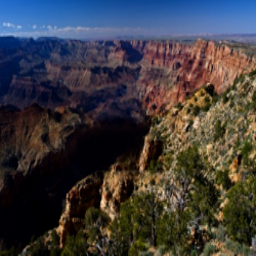}
	\end{subfigure}

	%%%%%%%%%%%%%%%
	%\setcounter{figure}{0}
	\caption{\label{fig:results_all_methods}\textbf{Qualitative comparison for arbitrary mask completion.} We  compare completions of a centered square of size $85\times85$ pixels.
	%We present the inpainting result of each method with a square centered mask of size of $85 \times 85$. One can clearly see where the mask is with 
	Shift-Map generates a completion with apparent boundaries, and DIP (both variants) and Patch-Match produce blurry outputs. In this specific scenery photo, our method performs at least on par with the external methods.
% 	CA, PIC and DSI.
	\vspace{-0.5mm}
	}
\end{figure*}{}

%% file: tables/realism_and_diversity.tex
       \begin{table*}[t]
            \begin{center}
            \smallskip\noindent
            \resizebox{1\linewidth}{!}{%
            %\raa{1.3}
            \begin{tabular}{@{}clclcccccccccccccccc@{}} 
                \toprule
                & Missing part & & & & \multicolumn{7}{c}{$85\times85$ centralized square} & & \multicolumn{7}{c}{$128\times128$ centralized square}\\
                \midrule
                & & & & & FID$\downarrow$ & & NIQE$\downarrow$ & & NIMA$\uparrow$ & &  LPIPS $\cdot10^{-3}$ & & FID$\downarrow$ & & NIQE$\downarrow$ & & NIMA$\uparrow$ & &  LPIPS $\cdot10^{-3}$\\
                & & & & & & & & & & &  (Diversity) & &  & & & &  & &  (Diversity)\\
                \midrule
                 & Non-Diverse & & CA & & 42.36 & & 4.35 & & 4.63 & & 0 & & 89.17 & & 4.39 & & 4.54 & & 0\\
                & & & LAMA & & 29.05 & & 4.58 & & 4.71 & & 0 & & 63.74 & & 4.58 & & 4.69 & & 0\\
                \cmidrule{2-20}
                &  & & DSI & & 38.01 & & 4.57 & & 4.58 & & $40$ & & 86.05 & & 4.57 & & 4.50 & & $110$\\
                External & & & PIC & & 42.86 & & 4.35 & & 4.69 & & $25$ & & 84.64 & & 4.11 & & 4.73 & & $86$\\
                & Diverse & & ICT & & 36.37 & & 4.25 & & 4.71 & & $38$ & & 76.85 & & 3.94 & & 4.70 & & $98$\\
                & & & CoMod-GAN & & 69.44 & & 4.13 & & 4.65 & & $15$ & & 102.40 & & 4.06 & & 4.65 & & $39$\\
                \midrule
                 & 
                 & & Shift-Map & & 45.26 & & 4.34 & & 4.66 & & 0 & & 100.9 & & 4.49 & & 4.54 & & 0\\
                % & & & DIP (optimized) & & 130.70 (85.97) & & 6.29 (5.80) & & 4.15 (4.30) & & 0 (0) & & 168.64 (129.52) & & 6.40 (5.85) & & 4.01 (4.05) & & 0 (0)\\
                % & & & PM & & 44.12 & & 4.58 & & 4.66 & & 0 & & 87.75 & & 4.79 & & 4.63 & & 0\\
                & Non-Diverse & & DIP optimized & & 85.97 & & 5.80 & &4.30 & & 0 & & 129.52 & & 5.85 & & 4.05 & & 0\\
                Internal & & & Patch-Match & & 44.12 & & 4.58 & & 4.66 & & 0 & & 87.75 & & 4.79 & & 4.63 & & 0\\
                \cmidrule{2-20}
                % & \multirow{2}{4em}{Diverse} & & PatchMatch & & 44.12 & & 4.58 & & 4.66 & & $12$ & & 87.75 & & 4.79 & & 4.63 & & $35$\\
                & Diverse & & IDC (Ours) & & 49.9 & & 4.56 & & 4.56 & & $8$ & & 97.85 & & 4.57 & & 4.47 & & $60$\\
                \bottomrule
            
            \end{tabular}}
        \end{center}
        \caption{\textbf{Quantitative evaluation.} We quantify visual quality and semantic diversity on the Places validation dataset. Our results are slightly inferior to external methods, which were trained specifically on this dataset, but are better than the optimized DIP (which outperforms the original DIP in all scores) and comparable to Shift-Map and Patch-Match. %Note however that Shift-Map often produces irregularities at the boundaries of the inpainted region and Patch-Match outputs blurry results (see Fig.~\ref{fig:results_all_methods}).
        \vspace{-0.5mm}}
        \label{table:realism_diversity}
        \end{table*}

%% file: sections/5_conclusions.tex
\section{Conclusions}

We presented a method for diverse image completion, %Our method is trained on a very limited amount of information- 
which does not require a training set. Since our model is trained only on the %the valid area of the input images. %on the
non-masked regions within the particular image the user wishes to edit, it is not constrained to any specific image domain or mask location or shape. Our approach can present to the user a variety of different completions with controllable diversity levels. %The internal ap
%This is the first work presenting internal diverse inpainting. We compared our method with non diverse and diverse methods and external and internal methods.
Extensive experiments show that our method is at least comparable to (and sometimes even better than) externally trained models in the task of object removal. %This is indicated by both quantitative measures and user studies. 
Furthermore, for completion of arbitrary regions, which may include parts of semantic objects, our method performs nearly as good as external methods. %The advantage of internal models is emphasized in images that are different from the dataset external models were trained on. In such cases, the statistics learned from the specific image will result in more realistic solutions compared to a model trained on a huge but different dataset.
%Our approach can handle any image size or aspect ratio, and is not limited to masks of specific sizes, location, or shapes.%Moreover, in our algorithm we offer the option to choose how diverse the user want the results.

% \input{figures/rare_domains}
% \input{figures/rare_domains2}

% \input{figures/fig1_more_results}

%% file: sections/6_supplementary.tex
\section{Training}
% \todo{optimizer, learning rates, run time, etc...}

\paragraph{Models and training time}
All the generators and discriminators are composed of 5 convolutional blocks consisting of \texttt{conv-BN-LeakyReLU}. The last block of the discriminator does not include an activation layer and the activation layer of the last block in the generator is Tanh instead of LeakyReLU. The slope of the Leaky-ReLU for negative values is 0.2. Our models (both generator and discriminator) consist of 32 channels in the first 4 scales. Every 4 scales the number of channels is increased by a factor of 2. At scales where the number of channels is modified (first and every 4th scale) we initialize the convolution weights and biases with normal distribution with mean=$0$ and std=$0.02$. %\tamar{is this correct?}. 
At all other scales, the weights and biases are initialized with those from the previous scale. Training takes about 2 hours on a Quadro RTX 8000 GPU for an image size of $193\times256$ and generating a new sample at inference takes less than a second per image.% \tomer{Say how many channels we use and at which scales this changes.}

\paragraph{Coarse scales}
At the coarse scales, the ``real'' image presented to the discriminator is a naively inpainted version of the masked input image, and therefore no masking is performed within the discriminator. Each of those coarse scales %a full real 
is trained for 2000 iterations using the Adam optimizer~\cite{kingma2014adam}. The learning rate for both the generator and the discriminator is $5\cdot 10^{-4}$ (which decreases after 1600 iterations by 0.1) and the momentum parameters are $\beta_1=0.5$ and $\beta_2=0.999$. Training is done with two losses: (i) WGAN-GP loss with gradient penalty weight of $\lambda=0.1$ (ii) Reconstruction loss with weight of $\alpha=10$. In each iteration we perform %the discriminator is updated over 
three gradient steps for updating $D$ followed by %and then the generator is updated over 
three gradient steps for updating $G$.

\paragraph{Fine scales}
For scales at which %When 
there exist enough valid patches, the ``real'' image presented to the discriminator is the masked input image, and masking is performed within the discriminator so as to ignore all patches containing masked pixels. Each such fine scale is trained for 3000 iterations using the Adam optimizer~\cite{kingma2014adam} with a learning rate of $5\cdot 10^{-5}$ for both the generator and discriminator, %is reduced to , 
with the same momentum parameters as the coarser scales. %of $\beta_1=0.5$ and $\beta_2=0.999$. 
%Let's denote $\delta$ to be the ratio between all pixels in the image to all valid pixels in the image and $\delta_{grad}$ to be the ratio between image size to the sum of all pixels in the gradual mask. 
We construct $\tilde{m}_n$, the soft version of the mask $m_n$ at the $n$-th scale, by dilating $m_n$ with a disc of size $\min(N-n,5)$ pixels and %is generated from 
convolving the result with a Gaussian filter with $\sigma=5$. The soft mask is then forced to be zeros at the invalid pixels of the image. Training is done with the same losses as in the coarse scales. We adjust the losses weights as follows: %with three losses: (i) Adversarial Loss 
(i) The gradient penalty weight is $\delta\cdot\lambda$ where $\delta$ is the ratio between the number of pixels within the image and the number of valid pixels and $\lambda=0.1$. (ii) The reconstruction loss weight is calculated according to $\delta_{rec}\cdot\alpha$ where $\delta_{rec}$ 
is the ratio between the number of pixels in the image and the sum of all elements in
% non-zero pixels in 
$\tilde{m}_n$. Here we take $\alpha=10$ for the coarser scales
% two thirds of the coarse scales 
and we increase it to $100$ in the last third of the scales. 
% coarse scales. %in the last third of the scales (total number of scales - coarse and fine) 
 %\tomer{Give details about the soft mask} 

\paragraph{Naive completion}
Obtaining the naive completion %inpainting 
is inspired by DIP~\cite{ulyanov2018deep}. %In order to get a real image for the coarse scales w
We train a U-Net network %\tamar{UNet?} 
with depth depending on the image size, for $1000$ iterations using the Adam optimizer with a learning rate of $10^{-3}$. Besides the losses we use, which are described in the main text, all the rest is done as in \cite{ulyanov2018deep}. 
% fixed input of Gaussian noise with $\mu=0$, $\sigma=1$ of size of the image to be inpainted $y_i$. 
% $y\times z$ \todo{add numbers}. 
% At each iteration we probate the input with an addative Gaussian noise with standard deviation of $0.03$. %is added to the input noise. 
% Training is done with two losses: (i) MSE between the output of the network and the real image in the valid pixels (ii) MSE between each inpainted pixel to its nearest neighbor in the valid pixels. \tamar{isn't this already mentioned in the paper?} \noa{Yes, I add it since we explain about the training but actually it is already in the paper} 
Examples of naive inpainting results are presented in Fig.~\ref{fig:init_inpainting}.

% \paragraph{}
\hfill \break
Code will be available upon acceptance.

%\tomer{Use different notations for figures in the supplementary. Call them Fig.~S1, Fig.~S2, etc. There's a way to change it automatically in Latex.}
\input{figures/Init_inpaint}
\clearpage
\section{Controlling the diversity}

% \paragraph{Controlling the diversity} 
For editing applications, users may wish to control the level of diversity. In our method there are two ways of controlling the diversity of the results: 
(i) by enlarging the standard deviation of the noise sampled during training, and
(ii) by enlarging the area of the new noise sampled at test time.

\paragraph{Enlarging the standard deviation of the injected noise}

We inject to the generator in all scales zero mean Gaussian noise with unit variance 1, which we multiply by a gain calculated according to the missing details in each scale. In order to check the influence of the standard deviation on the diversity of our method, we trained three different models on the same image. Each model was trained with a different noise variance (1,2, and 4).
For each model, we calculated the semantic diversity (LPIPS) over 1000 pairs of samples. As can be seen in Tab.~\ref{table:noise_var}, enlarging the variance of the noise has a relatively small effect on the diversity of our method.
\input{tables/noise_std}
\paragraph{Enlarging the area of the new noise sampled}
An alternative way of controlling diversity is by changing the area of the new noise sampled at test time. Sampling a larger area in each pyramid scale results in higher diversity, but may also modify the valid area of the image. %Thus we let the user choose how much diversity is desired. %wanted versus preserving all the known pixels in the 
This trade-off is demonstrated in Fig.~\ref{fig:diversity_examples}. %In fact, we offer the user two additional operation modes. %(i) Low diversity: this approach will be used when we only want to fill the missing pixels without changing the area outside the mask. In order to do that we erode half a receptive field (including multi-scale upsampling RF) from the mask $\Mask_n$ at each scale as seen in fig.~\ref{fig:diverse_methods}. In this approach the main diversity between different solutions will be only inside the mask \Mask. This approach is presented in fig.~\ref{fig:diversity_examples} as IDC. Note that there still might be small changes outside the mask, however it is negligible. \tamar{I think we should also refer to a figure with results, not only the masks.} In this approach when the initial mask is relatively small, after erosion the mask might be all zeros in the coarser scales, leading to low diversity.
Here, high diversity refers to the setting where the new noise fills the entire mask area, without erosion. %in the given masks [$m_N,...,m_0$], and change all pixels on the real image $\real_0$ that were affected from the noise sampled $\noise_n$ in all scales as described as IDC$^{high}$ in  fig.~\ref{fig:diversity_examples}.
Medium diversity refers to %user can choose intermediate diversity %between the original IDC and between those two approaches, between strictly preserving the knowing pixels or enlarging to the maximum the diversity, one can choose an intermediate approach and 
partially eroding the mask (by less than half a receptive-field). %and get higher diversity than the first approach without preserving some known pixels around the missing pixels. This in between approaches (medium diversity) is presented as IDC$^{mid}$ in fig.~\ref{fig:diversity_examples}.
%Visual examples for different diversity modes are presented in fig.~\ref{fig:diversity_examples}, along with diversity maps quantifying the variability of the different completions at each operation mode. 
Note that in our normal diversity setting, we intentionally do not erode half of the full receptive field (which includes the up-sampling operation), in order to allow increased diversity. This results in only minor changes to the regions outside the mask (see STD image). The semantic diversity of our method with these three modes is shown in Tab.~\ref{table:diversity_comparison}. %Indeed, the score increases as we allow larger portion of $\noise^\text{test}_n$ to be resampled at inference. }

\input{figures/diversity}
\input{tables/diversity}

\clearpage

\section{Additional results}
Figure~\ref{fig:figure1_places} shows inpainting results for the image presented in %same images as 
Fig.~1 in the main text, %paper, 
where here we use the DSI~\cite{peng2021generating} and ICT~\cite{wan2021high} models that were trained on the Places dataset, rather than the ones trained on Imagenet (CoMod-GAN does not have an ImageNet model). One can see that with these models as well, DSI and ICT introduce %results have 
artifacts. This is while our results %and we generate 
are more realistic. Additional examples for our diverse completions %for our method 
are presented in Fig.~\ref{fig:more_results}.
Figure~\ref{fig:qualitative_comparison_SM} presents the results of other models, for the images presented in Fig.~6 in the main text.
\input{figures/figure1_places}

\input{figures/more_results}
\input{figures/partImagenet_nondiverse_latex_SM}
% \clearpage
\section{Importance of the Coarse Scales}
The first row of Fig.~\ref{fig:no_naive_inpainting} shows the results of our method when omitting %not training 
the coarser scales in the model. Namely, %meaning that 
we start the training from the first masked scale, $i-1$. For comparison, the last row shows %represent 
our full training scheme. As can be seen, the coarse scales are crucial for preserving global structures within the image.
\input{figures/no_naive_inpainting}

\section{Importance of Masking the BN Layer}
At the fine scales, where we do perform masking, %the real image is a masked image, 
the discriminator needs to ignore all invalid pixels in the real image. This is done by masking the elements of the discrimination map that correspond to patches containing invalid pixels. %However, it is not enough to ignore patches affected by invalid pixels at the output of the discriminator, 
In a convolutional model without BN layers, this operation guarantees that the discriminator is not affected by the missing region. However, in the presence of BN layers, this is not enough because the BN statistics (mean and variance) are computed over their entire input feature map. Therefore, it is also crucial to compute the BN statistics in each layer only over valid features (not affected by invalid pixels). Figure~\ref{fig:no_D_masking} illustrates the importance of passing through the BN layer only valid patches. 
% \tamar{isn't the last row in the figure and in the previous the same? maybe conmine them to one figure with three rows?} \noa{I added a third figure which combine both figures (left the other ones to see what we choose)}
One can see that when not masking the BN layer, the results are blurry and contain artifacts. %\tomer{Start with a sentence that reminds the reader what you're talking about.}
\input{figures/no_D_masking}
% \input{figures/no_naive_and_masking}
% \clearpage
\section{Limitation}
In the setting of inpainting, the mask may hide parts of semantic objects, which internal methods cannot complete, as presented in Fig.~\ref{fig:limitation}. However, this task is challenging even for external methods.
\input{figures/limitation}

\section{Patch-Match} A comment is in place regarding Patch-Match, which uses a stochastic algorithm for finding approximate nearest neighbor patches. While this algorithm should theoretically converge to the same solution despite its randomness,
% without being affected by 
%the initialization. 
% of k-nearest neighbors. However 
in practice 
% this is not the case, and 
different runs
% seeds in PM output 
result in slightly different completions. 
% solutions. 
Superficially, this could suggest that Patch-Match should also be regarded as an internal diverse method. However, in practice, the diversity arising from Patch-Match's randomness is actually quite low. For example, for $128\times 128$ masks, Patch-Match's LPIPS diversity is $35\times10^{-3}$ while ours is $60\times10^{-3}$. We therefore do not regard this unintentional byproduct of the algorithm's implementation as genuine diversity. %\tamar{maybe move this paragraph to SM? with the subsections now it's strange to have it here. And actually I'm not sure we need it in the main text.}
% For smaller masks, PM's diversity is similar to ours (\eg for $85\times 85$, PM's diversity is $12\times10^{-3}$ and ours is $8\times10^{-3}$). 
%However PM's visual quality is inferior to ours as it tends to generate blurry and repetitive completions.

\clearpage
\section{Completion of images from other domains}
Figure~\ref{fig:rare_images_SM} shows completion results for the images of Fig.~2 in the main text, obtained using competing methods. One can see that external methods often insert objects to the images.
\input{figures/rare_images_SM}

\clearpage
% \section{Variance of injected noise}

\section{Modified DIP}
Recall that in the coarse scales of our model, we use naive inpainting, which we obtain from a modified variant of the DIP method. Figure~\ref{fig:modified_dip} shows comparisons between the original DIP algorithm and our modified DIP. One can see the importance of adding the color consistency loss into to original DIP. Our modified DIP scheme was optimized for low-resolution images (we only used it for the coarse pyramid scales). Using it `as is' on the full resolution images does not improve upon the original DIP.
\input{figures/modified_dip}

\clearpage

\section{AMT survey}
%In the paper we present the results of the user study done in order to compare our method with baselines. 
We start the survey with 5 tutorial questions with feedback. %Each user was trained with a tutorial of 5 images, 
In the tutorial questions, it is clear which completion is better. Figure~\ref{fig:tutorial} depicts the images that participated in the tutorial.  % images. 
After the tutorial, the user was presented with %to 
45 questions. In each of them, %task 
we display %show 
the original image, the masked image, our completion and one competing completion. All images presented in the user study are shown in Fig~\ref{fig:user_study}.
\input{figures/tutorial}
\input{figures/user_study}
\clearpage
\section{Places validation images}

For comparison to baselines on the task of inpainting with arbitrary
masks (not necessarily hiding a whole semantic object), we used 50 images from the Places dataset, presented in Fig.~\ref{fig:palces2}.
\input{figures/places2}